\documentclass[10pt,journal,compsoc]{IEEEtran}

\usepackage{amssymb}
\usepackage{amsmath}
\usepackage{amsthm}
\usepackage{bbm}
\usepackage{booktabs} 
\usepackage{chemfig}
\usepackage{graphicx}
\usepackage{caption}
\usepackage{cite}
\usepackage{mathtools}

\usepackage{url}
\usepackage{multirow}
\usepackage{float}
\usepackage[ruled, vlined,algo2e,linesnumbered]{algorithm2e}
\usepackage{tikz}
\usepackage{xcolor}
\usepackage{tkz-euclide,subfigure}
\usetikzlibrary{shapes}
\newlength{\R}
\usepackage{pgfplots}
\pgfplotsset{compat=newest}
\usetikzlibrary{shapes.arrows}
\usetikzlibrary{arrows.meta}
\usetikzlibrary{trees,arrows}

\makeatletter
\newcommand{\nosemic}{\renewcommand{\@endalgocfline}{\relax}}
\newcommand{\dosemic}{\renewcommand{\@endalgocfline}{\algocf@endline}}
\newcommand{\comment}[1]{}

\newcommand{\matrixSym}[1]{\ensuremath{\mathbf{\uppercase{#1}}}}

\newcommand\footnoteref[1]{\protected@xdef\@thefnmark{\ref{#1}}\@footnotemark}
\newcommand{\RN}[1]{%
	\textup{\uppercase\expandafter{\romannumeral#1}}%
}
\DeclareMathAlphabet{\mathcal}{OMS}{cmsy}{m}{n}

\newtheorem{theorem}{Theorem}
\newtheorem{definition}{Definition}

\usepackage{flushend}
\usepackage{balance}
\begin{document}
	
\title{Learning Deep Graph Representations via Convolutional Neural Networks}

\author{Wei~Ye, Omid~Askarisichani, Alex~Jones, and Ambuj~Singh
	\IEEEcompsocitemizethanks{\IEEEcompsocthanksitem W. Ye, O. Askarisichani, A. Jones, and A. Singh are with the Department of Computer Science, University of California, Santa Barbara.\protect\\ Email: \{weiye, omid55, alexjones, ambuj\}@cs.ucsb.edu}
	\thanks{Manuscript received xxx; revised xxx.}}

\markboth{IEEE TRANSACTIONS ON KNOWLEDGE AND DATA ENGINEERING}%
{Ye \MakeLowercase{\textit{et al.}}: Learning Deep Graph Representations via Convolutional Neural Networks}

\IEEEtitleabstractindextext{%
	\begin{abstract}
		Graph-structured data arise in many scenarios. A fundamental problem is to quantify the similarities of graphs for tasks such as classification. R-convolution graph kernels are positive-semidefinite functions that decompose graphs into substructures and compare them. One problem in the effective implementation of this idea is that the substructures are not independent, which leads to high-dimensional feature space. In addition, graph kernels cannot capture the high-order complex interactions between vertices. To mitigate these two problems, we propose a framework called \textsc{DeepMap} to learn deep representations for graph feature maps. The learned deep representation for a graph is a dense and low-dimensional vector that captures complex high-order interactions in a vertex neighborhood. \textsc{DeepMap} extends Convolutional Neural Networks (CNNs) to arbitrary graphs by generating aligned vertex sequences and building the receptive field for each vertex. We empirically validate \textsc{DeepMap} on various graph classification benchmarks and demonstrate that it achieves state-of-the-art performance.
	\end{abstract}
	
	}

\maketitle

\IEEEdisplaynontitleabstractindextext

\IEEEpeerreviewmaketitle

\section{Introduction}\label{intro}
Irregular data arise in many scenarios, such as proteins or molecules in bioinformatics, communities in social networks, text documents in natural language processing, and images annotated with semantics in computer vision. Graphs are naturally used to represent such data. One fundamental problem with graph-structured data is computing their similarities, needed for downstream tasks such as classification. Graph kernels have been developed and widely used to measure the similarities between graph-structured data. This paper deals with graph kernels that are instances of the family of R-convolution kernels~\cite{haussler1999convolution}. The key idea is to recursively decompose graphs into their substructures such as graphlets~\cite{prvzulj2004modeling}, subtrees ~\cite{shervashidze2009fast, shervashidze2011weisfeiler}, walks~\cite{vishwanathan2010graph,zhang2018retgk}, paths ~\cite{borgwardt2005shortest,ye2019tree++}, and then compare these substructures from two graphs. A typical definition for graph kernels is $\mathcal{K}(\mathcal{G}_1, \mathcal{G}_2) = \left\langle \phi(\mathcal{G}_1), \phi(\mathcal{G}_2)\right\rangle$, where $\left\langle \cdot, \cdot\right\rangle $ denotes the dot product between two vectors, $\phi(\mathcal{G}_i)=\left[ \psi(\mathcal{G}_i,A_1),\psi(\mathcal{G}_i,A_2),\ldots,\psi(\mathcal{G}_i,A_m)\right]$ is a vector that contains the number of occurrences of substructure $A_j$ ($1\leq j\leq m$) (denoted as $\psi(\mathcal{G}_i,A_j)$) in graph $\mathcal{G}_i$ ($i=1,2$). We call $\phi(\mathcal{G})$ the feature map\footnote{In this work, feature map and representation are used in an exchangable manner.} (please see Definition~\ref{def:fmg}) of graph $\mathcal{G}$.

\begin{figure}[!htb]
	\hspace*{\fill}
	\centering
	\subfigure[$G_1^{(3)}$]{
		\begin{tikzpicture}
		\begin{scope}[scale=1.2,every node/.style={circle,draw}]
		\node (A) at (0,0) {};
		\node (B) at (1.5,0) {};
		\node (C) at (0.75,1.3) {};
		\end{scope}
		\end{tikzpicture}
	}
	\hfill
	\centering
	\subfigure[$G_2^{(3)}$]{
		\begin{tikzpicture}
		\begin{scope}[scale=1.2,every node/.style={circle,draw}]
		\node (A) at (0,0) {};
		\node (B) at (1.5,0) {};
		\node (C) at (0.75,1.3) {};
		\end{scope}
		
		\begin{scope}[>={Stealth[black]},
		every node/.style={fill=white,circle},
		every edge/.style={draw=red,very thick}]
		\foreach \from/\to in {A/B}
		\draw[-] (\from) -- (\to);
		\end{scope}
		\end{tikzpicture}
	}
	\hspace*{\fill}
	
	\hspace*{\fill}
	\centering
	\subfigure[$G_3^{(3)}$]{
		\begin{tikzpicture}
		\begin{scope}[scale=1.2,every node/.style={circle,draw}]
		\node (A) at (0,0.65) {};
		\node (B) at (1.5,0.65) {};
		\node (C) at (0.75,0.65) {};
		\end{scope}
		
		\begin{scope}[>={Stealth[black]},
		every node/.style={fill=white,circle},
		every edge/.style={draw=red,very thick}]
		\foreach \from/\to in {A/C,B/C}
		\draw[-] (\from) -- (\to);
		\end{scope}
		\end{tikzpicture}
	}
	\hfill
	\centering
	\subfigure[$G_4^{(3)}$]{
		\begin{tikzpicture}
		\begin{scope}[scale=1.2,every node/.style={circle,draw}]
		\node (A) at (0,0) {};
		\node (B) at (1.5,0) {};
		\node (C) at (0.75,1.3) {};
		\end{scope}
		
		\begin{scope}[>={Stealth[black]},
		every node/.style={fill=white,circle},
		every edge/.style={draw=red,very thick}]
		\foreach \from/\to in {A/B,A/C,B/C}
		\draw[-] (\from) -- (\to);
		\end{scope}
		\end{tikzpicture}
	}
	\hspace*{\fill}
	\caption{Non-isomorphic subgraphs (graphlets) of size $k=3$.}
	\label{fig:graphlets}
\end{figure}
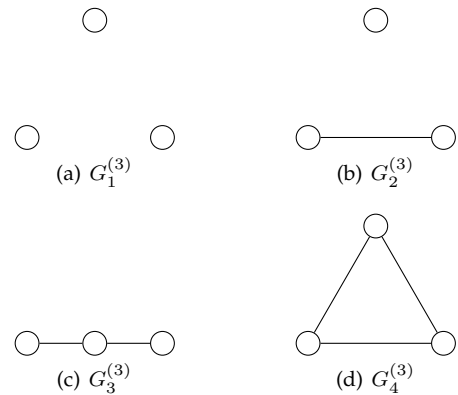

Although graph kernels are efficient methods to compute graph similarities, they still have the following two main issues: First, the substructures extracted from graphs are not independent. For instance, by adding/deleting vertices or edges, one graphlet can be derived from another graphlet. Figure~\ref{fig:graphlets} shows that graphlet $G_3^{(3)}$ can be derived from graphlet $G_2^{(3)}$ by adding an edge. This dependency (redundancy) remains in graph feature maps. Because of this dependency between substructures, the dimension of the graph feature map often grows exponentially and thus it leads to low effectiveness. Second, graph kernels use the hand-crafted features without considering the complex interactions between vertices. Thus, high-order information in the neighborhood of a vertex is not integrated into graph feature maps.
%

To solve the first main issue, Deep Graph Kernels (DGK)~\cite{yanardag2015deep} leverages techniques from natural language processing to learn latent representations for substructures. Then the similarity matrix between substructures is computed and integrated into the computation of the graph kernel matrix. If the number of substructures is high, it will cost a lot of time and memory to compute the similarity matrix. In addition, DGK uses natural language processing models to learn latent representations for substructures without proving that the frequency of substructures extracted from graphs follows a power-law distribution, which is observed in natural language. For example, the Weisfeiler-Lehman subtree kernel (WL)~\cite{shervashidze2009fast, shervashidze2011weisfeiler} decomposes graphs into subtree patterns and then counts the number of common subtree patterns across graphs. If the subtree is of depth zero (i.e., only one root vertex), we can represent it using the vertex degree. However, the vertex degree distribution of a graph does not always follow a power-law distribution. Thus, the learned representations for substructures are not accurate. To solve the second main issue, people develop graph neural networks (GNNs)~\cite{xu2018powerful,niepert2016learning,zhang2018end,duvenaud2015convolutional} to extract the complex high-order interactions in a vertex neighborhood. 


 To mitigate these two main issues, we develop a CNN architecture on the vertex feature map (please see Definition~\ref{def:fmv}) extracted from each vertex in a graph. The method is called \textsc{DeepMap} since it learns deep representations for graph feature maps. The learned deep representation of a graph is a dense and low-dimensional vector that captures complex high-order interactions in a vertex neighborhood. Typically, a CNN contains several convolutional and dense layers. CNNs exploit spatial locality of an input and thus the learned ``filters'' produce the strongest response to the spatially local input pattern. Stacking many such layers leads to non-linear filters that can capture appropriate patterns. The extension of CNNs from images to graphs of arbitrary size and shape faces one main challenge: as opposed to images whose pixels are spatially ordered, vertices in graphs do not have spatial or temporal order. Vertices across different graphs are hard to align, and thus the receptive fields of CNNs cannot be directly applied on vertices in graphs. To develop a CNN applicable to arbitrary graphs, we propose to solve two main problems: (1) Generate a vertex sequence for each graph such that these sequences are aligned. (2) Determine the receptive field for each vertex in each vertex sequence.

Our contributions are summarized as follows: 
\begin{itemize}
	\item We analyze the graph feature maps of three popular graph kernels and then propose the definition of vertex feature maps.
	\item We develop a new CNN model \textsc{DeepMap} on the vertex feature maps to mitigate the two main issues. The extension of CNN from images to graphs of arbitrary size and shape is achieved by two steps: (1) We use eigenvector centrality~\cite{bonacich1987power} as a measure to generate aligned vertex sequences. (2) We use a breadth-first search (BFS) method for constructing the receptive field for each vertex in each vertex sequence.
	\item We empirically validate \textsc{DeepMap} on a number of graph classification benchmarks and demonstrate that it achieves state-of-the-art performance.
\end{itemize} 

The rest of the paper is organized as follows: We describe related work in Section 2. Section 3 covers the ideas of graph feature maps of three popular graph kernels. Section 4 introduces the core ideas behind our approach \textsc{DeepMap}, including the definition of vertex feature maps and the extension of CNN to arbitrary graphs. Using the benchmark graph datasets, Section 5 compares \textsc{DeepMap} with related techniques. Section 6 makes some discussions. And Section 7 concludes the paper.

\section{Related Work}\label{relatedWork}
\subsection{Graph Kernels}
R-convolution graph kernels can be based on walks~\cite{vishwanathan2010graph,zhang2018retgk}, paths~\cite{borgwardt2005shortest,ye2019tree++}, graphlets~\cite{shervashidze2009efficient},  and subtree patterns~\cite{ramon2003expressivity,mahe2009graph,shervashidze2009fast, shervashidze2011weisfeiler}, etc. RetGK~\cite{zhang2018retgk} introduces a structural role descriptor for vertices, i.e., the return probabilities features (RPF) generated by random walks. The RPF is then embedded into the Hilbert space where the corresponding graph kernels are derived. The shortest-path graph kernel (SP)~\cite{borgwardt2005shortest} counts the number of pairs of shortest paths that have the same source and sink labels and the same length in two graphs. The Tree++~\cite{ye2019tree++} graph kernel is proposed for the problem of comparing graphs at multiple levels of granularities. It first uses a path-pattern graph kernel to build a truncated BFS tree rooted at each vertex and then uses paths from the root to every vertex in the truncated BFS tree as features to represent graphs. To capture graph similarity at multiple levels of granularities, Tree++ incorporates a new concept called super path into the path-pattern graph kernel. The super path contains truncated BFS trees rooted at the vertices in a path. The graphlet kernel (GK)~\cite{shervashidze2009efficient} proposes to use the method of random sampling to extract graphlets from graphs. The idea of random sampling is motivated by the observation that the more sufficient number of random samples is drawn, the closer the empirical distribution to the actual distribution of graphlets in a graph. The Weisfeiler-Lehman subtree kernel (WL)~\cite{shervashidze2009fast, shervashidze2011weisfeiler} is based on the Weisfeiler-Lehman test of graph isomorphism~\cite{weisfeiler1968reduction} for graphs. In each iteration, the Weisfeiler-Lehman test of graph isomorphism augments vertex labels by concatenating their neighbors' labels and then compressing the augmented labels into new labels. The compressed labels correspond to the subtree patterns. WL counts common original and compressed labels in two graphs.

There are also some graph kernels~\cite{bai2015quantum,johansson2014global} focusing on computing global similarities between graphs. The paper~\cite{bai2015quantum} computes the Jensen-Shannon divergence between probability distributions over graphs, without the need of decomposing the graph into substructures. The paper~\cite{johansson2014global} designs two novel graph kernels to capture global properties of unlabeled graphs. The kernels are based on the Lov\'{a}sz number and are called the Lov\'{a}sz $\vartheta$ kernel and the SVM-$\vartheta$ kernel. Both of these two kernels still need to enumerate all subsets of nodes from two graphs and compute the Lov\'{a}sz number for each subset. The number of possible subsets of nodes still exponentially increases with the increasing size of the graph.


Recently, some research works such as~\cite{yanardag2015deep, kriege2016valid} focus on augmenting the existing graph kernels or fusing GNNs with graph kernels~\cite{du2019gntk}. DGK~\cite{yanardag2015deep} deals with the problem of diagonal dominance in graph kernels. The diagonal dominance means that a graph is more similar to itself than to any other graphs in the dataset because of the sparsity of common substructures across different graphs. DGK leverages techniques from natural language processing to learn latent representations for substructures. Then the similarity matrix between substructures is computed and integrated into graph kernels. If the number of substructures is high, it will cost a lot of time and memory to compute the similarity matrix. OA~\cite{kriege2016valid} develops some base kernels that generate hierarchies from which the optimal assignment kernels are computed. The optimal assignment kernels can provide a more valid notion of graph similarity. The authors finally integrate the optimal assignment kernels into the Weisfeiler-Lehman subtree kernel. Graph Neural Tangent Kernel (GNTK)~\cite{du2019gntk} is inspired by the connections between over-parameterized neural networks and kernel methods~\cite{jacot2018neural,arora2019exact}. It is a model that inherits both advantages from GNNs  and graph kernels. It can extract powerful features from graphs as GNNs and is easy to train and analyze as graph kernels. It is equivalent to infinitely wide GNNs trained by gradient descent.

\subsection{Graph Neural Networks}
In addition to the above-described literature, there are also some literature from the field of graph neural networks (GNNs)~\cite{xu2018powerful,niepert2016learning,atwood2016diffusion,zhang2018end,duvenaud2015convolutional,henaff2015deep,kipf2016semi,defferrard2016convolutional,velivckovic2017graph} related to our work. SpectralNet~\cite{henaff2015deep} develops an extension of spectral networks~\cite{bruna2013spectral} for deep learning on graphs. A spectral network generalizes a convolutional network through the Graph Fourier Transform. Graph-CNN~\cite{defferrard2016convolutional} proposes a strictly localized spectral filters that uses Chebyshev polynomials for approximately learning K-order spectral graph convolutions~\cite{hammond2011wavelets}. Both SpectralNet and Graph-CNN first construct similarity graphs from a dataset and then classify data points into different classes. They are not applicable to graphs of arbitrary size and shape. GCN~\cite{kipf2016semi} introduces a simple and well-behaved layer-wise propagation rule for graph convolutional networks. The propagation rule is derived from the first-order approximation of spectral graph convolutions. GAT~\cite{velivckovic2017graph} computes the latent representations for each vertex in a graph, by attending over its neighbors, following a self-attention strategy. It specifies different weights to different vertices in a neighborhood. GraphSAGE~\cite{hamilton2017inductive} is developed for the inductive representation learning on graphs. It learns a function to generate embeddings for each node, by sampling and aggregating features from a node's local neighborhood. GCN, GAT and GraphSAGE are designed for the classification of vertices in a graph.

Neural Graph Fingerprints (NGF)~\cite{duvenaud2015convolutional} introduces a convolutional neural network on graphs for learning differentiable molecular fingerprints, by replacing each discrete operation in circular fingerprints with a differentiable analog. NGF develops a local message-passing architecture that propagates information to a depth of $R$ neighborhood. \textsc{DCNN}~\cite{atwood2016diffusion} extends convolutional neural networks to graphs by introducing a diffusion-convolution operation, based on which diffusion-based representations can be learned from graphs and used as an effective basis for vertex classification and graph classification. \textsc{DGCNN}~\cite{zhang2018end} first designs a novel special graph convolution layer to extract multi-scale vertex features. Then, in order to sequentially read graphs of differing vertex orders, \textsc{DGCNN} designs a novel SortPooling layer that sorts graph vertices in a consistent order so that traditional neural networks can be trained on graphs. \textsc{GIN}~\cite{xu2018powerful} is proposed to analyze the expressive power of GNNs to capture different graph structures. Both \textsc{DGCNN} and \textsc{GIN} are inspired by the close connection between GNNs and the Weisfeiler-Lehman test of graph isomorphism. The inputs to \textsc{DGCNN} and \textsc{GIN}  are the one-hot encodings of vertex labels.
\textsc{PatchySan}~\cite{niepert2016learning} generalizes CNNs from images to arbitrary graphs. It first orders vertices by the graph canonicalization tool \textsc{Nauty}~\cite{mckay2014practical}, and then performs three operations: (1) vertex sequence selection, (2) neighborhood assembly, and (3) graph normalization. There is another work that also uses CNNs on graphs. DeepTrend 2.0~\cite{dai2019deeptrend} proposes a CNN-based model on a sensor network for traffic flow prediction. It converts the sensor network into an image, in which neighboring pixels represent sensors that have a strong correlation. In this way, the local similarity of the image is fullfilled. But neighboring sensors in a sensor network may not be mapped to the neighboring pixels in the image. 

\section{Graph Feature Maps} \label{pre}
In this work, we use lower-case Roman letters (e.g.,\ $a,b$) to denote scalars. We denote vectors (row) by boldface lower case letters (e.g.,\ $\mathbf{x}$) and denote its $i$-th element by $\mathbf{x}(i)$. We use $\mathbf{x}=[x_1,\ldots, x_n]$ to denote creating a vector by stacking scalar $x_i$ along the columns. We consider an undirected labeled graph $\mathcal{G}=(\mathcal{V},\mathcal{E}, l)$, where $\mathcal{V}$ is a set of graph vertices with number $|\mathcal{V}|$ of vertices, $\mathcal{E}$ is a set of graph edges with number $|\mathcal{E}|$ of edges, and $l: \mathcal{V}\rightarrow \Sigma$ is a function that assigns labels from a set of positive integers $\Sigma$ to vertices. Without loss of generality, $\lvert\Sigma\rvert\leq |\mathcal{V}|$. An edge $e$ is denoted by two vertices $uv$ that are connected to it. In graph theory~\cite{harary1969graph}, a walk is defined as a sequence of vertices, e.g., $\left( v_1,v_2,\ldots\right) $ where consecutive vertices are connected by an edge. A trail is a walk that consists of all distinct edges. A path is a trail that consists of all distinct vertices and edges. The depth of a subtree is the maximum length of paths between the root and any other vertex in the subtree. 

\begin{definition}[Graph Isomorphism]
	\label{def:gi}
	Two undirected labeled graphs $\mathcal{G}_1=(\mathcal{V}_1,\mathcal{E}_1, l_1)$ and $\mathcal{G}_2=(\mathcal{V}_2,\mathcal{E}_2, l_2)$ are isomorphic (denoted by $\mathcal{G}_1 \simeq \mathcal{G}_2$) if there is a bijection $\varphi: \mathcal{V}_1 \rightarrow \mathcal{V}_2$, (1) such that for any two vertices $u, v \in \mathcal{V}_1$, there is an edge $uv$ if and only if there is an edge $\varphi(u)\varphi(v)$ in $\mathcal{G}_2$; (2) and such that $ l_1(v) = l_2(\varphi(v))$.
\end{definition}

Let $\mathcal{X}$ be a non-empty set and let $\mathcal{K}: \mathcal{X} \times \mathcal{X} \rightarrow \mathbb{R}$ be a function on $\mathcal{X}$. Then $\mathcal{K}$ is a kernel on $\mathcal{X}$ if there is a real Hilbert space $\mathcal{H}$ and a mapping $\phi:  \mathcal{X} \rightarrow \mathcal{H}$ such that $\mathcal{K}(x, y) = \langle\phi(x), \phi(y)\rangle$ for all $x$, $y$ in $\mathcal{X}$, where $\langle\cdot, \cdot\rangle$ denotes the inner product of $\mathcal{H}$, $\phi$ is called a feature map and $\mathcal{H}$ is called a feature space. $\mathcal{K}$ is symmetric and positive-semidefinite. In the case of graphs, let $\phi(\mathcal{G})$ denote a mapping from a graph to a vector which contains the number of occurrences of the atomic substructures in graph $\mathcal{G}$. Then, the kernel on two graphs $\mathcal{G}_1$ and $\mathcal{G}_2$ is defined as $\mathcal{K}(\mathcal{G}_1, \mathcal{G}_2) = \langle\phi(\mathcal{G}_1), \phi(\mathcal{G}_2)\rangle$.

We define graph feature maps as follows:
\begin{definition}[Graph Feature Maps]
	\label{def:fmg}
	Define a map $\psi: \{\mathcal{G}_1,\mathcal{G}_2,\ldots,\mathcal{G}_n\} \times \Sigma\rightarrow \mathbb{N}$ such that $\psi(\mathcal{G}, A)$ is the number of occurrences of the atomic substructure $A$ in graph $\mathcal{G}$. Then the feature map of graph $\mathcal{G}$ is defined as follows:
	\begin{equation}
	\label{equ:graph}
	\phi(\mathcal{G})=\left[ \psi(\mathcal{G},A_1),\psi(\mathcal{G},A_2),\ldots,\psi(\mathcal{G},A_m)\right]
	\end{equation}
	where $m$ is the number of unique atomic substructures and depends on graphs.
\end{definition}

In the following, let us elaborate the mechanisms of three popular graph kernels, i.e., the graphlet kernel (GK)~\cite{shervashidze2009efficient}, the shortest-path kernel (SP)~\cite{borgwardt2005shortest} and the Weisfeiler-Lehman subtree kernel (WL)~\cite{shervashidze2011weisfeiler}, and relate them to our definitions.

A graphlet $G$ (as shown in Figure~\ref{fig:graphlets}) is a non-isomorphic subgraph of size $k$ induced from graph $\mathcal{G}$. Let $\mathsf{G}^{(k)}$ be the multiset\footnote{A set that can contain the same element multiple times.} of size-$k$ graphlets. Then, for graph $\mathcal{G}$, its feature map is defined as follows:
\begin{equation}
\label{equ:graphlet}
\phi(\mathcal{G})=\left[ \psi(\mathcal{G},G_1^{(k)}),\psi(\mathcal{G},G_2^{(k)}),\ldots,\psi(\mathcal{G},G_m^{(k)})\right]
\end{equation}
where $m$ stands for the number of unique graphlets of size $k$ in $\mathsf{G}^{(k)}$, $\psi(\mathcal{G},G_i^{(k)}) (1\leq i\leq m)$ denotes the frequency of the unique graphlet $G_i^{(k)}$ occuring in graph $\mathcal{G}$. Exhaustive enumation of all graphlets of size $k$ is prohibitively expensive, especially for large graphs. Usually, we use some sampling techniques such as the random sampling scheme proposed in~\cite{shervashidze2009efficient} to sample a number of $q$ graphlets of size $k$ from graph $\mathcal{G}$, and then count the frequency of each unique graphlet occurring in these $q$ samples.

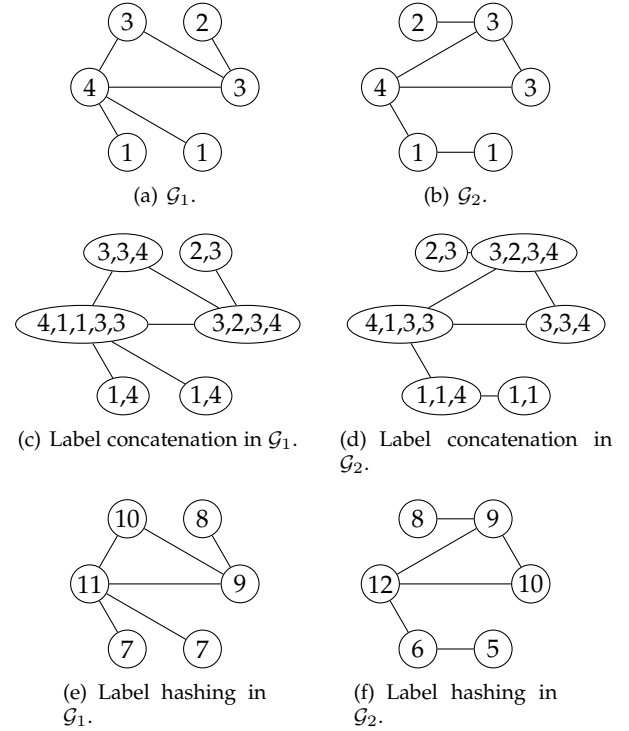
\begin{figure}[htb]
	\centering
	\hspace*{\fill}
	\subfigure[$\mathcal{G}_1$.]{
		\begin{tikzpicture}
		[scale=1,every node/.style={draw, circle, minimum size=1.5em,inner sep=1}]
		\node (n6) at (60: \R) {2};
		\node (n4) at (120: \R)  {3};
		\node (n5) at (180: \R)  {4};
		\node (n1) at (240: \R) {1};
		\node (n2) at (300: \R)  {1};
		\node (n3) at (360: \R)  {3};
		
		\foreach \from/\to in {n4/n5,n5/n1,n2/n5,n3/n4,n5/n3,n6/n3}
		\draw[] (\from) -- (\to);
		\end{tikzpicture}
	}
	\hfill
	\centering
	\subfigure[$\mathcal{G}_2$.]{
		\begin{tikzpicture}
		[scale=1,every node/.style={draw, circle, minimum size=1.5em,inner sep=1}]
		\node (n6) at (60: \R) {3};
		\node (n4) at (120: \R)  {2};
		\node (n5) at (180: \R)  {4};
		\node (n1) at (240: \R) {1};
		\node (n2) at (300: \R)  {1};
		\node (n3) at (360: \R)  {3};
		
		\foreach \from/\to in {n6/n4,n5/n1,n2/n1,n5/n6,n5/n3,n6/n3}
		\draw[] (\from) -- (\to);
		\end{tikzpicture}
	}
	\hspace*{\fill}
	
	\centering
	\hspace*{\fill}
	\subfigure[Label concatenation in $\mathcal{G}_1$.]{
		\begin{tikzpicture}
		[scale=1.1,every node/.style={draw, ellipse}]
		\node[inner sep=1pt,minimum size=4pt] (n6) at (60: \R) {2,3};
		\node[inner sep=1pt,minimum size=4pt] (n4) at (120: \R)  {3,3,4};
		\node[inner sep=1pt,minimum size=4pt] (n5) at (180: \R)  {4,1,1,3,3};
		\node[inner sep=1pt,minimum size=4pt] (n1) at (240: \R) {1,4};
		\node[inner sep=1pt,minimum size=4pt] (n2) at (300: \R)  {1,4};
		\node[inner sep=1pt,minimum size=4pt] (n3) at (360: \R)  {3,2,3,4};
		
		\foreach \from/\to in {n4/n5,n5/n1,n2/n5,n3/n4,n5/n3,n6/n3}
		\draw[] (\from) -- (\to);
		\end{tikzpicture}
	}
	\hfill
	\centering
	\subfigure[Label concatenation in $\mathcal{G}_2$.]{
		\begin{tikzpicture}
		[scale=1.1,every node/.style={draw, ellipse}]
		\node[inner sep=1pt,minimum size=4pt] (n6) at (60: \R) {3,2,3,4};
		\node[inner sep=1pt,minimum size=4pt] (n4) at (120: \R)  {2,3};
		\node[inner sep=1pt,minimum size=4pt] (n5) at (180: \R)  {4,1,3,3};
		\node[inner sep=1pt,minimum size=4pt] (n1) at (240: \R) {1,1,4};
		\node[inner sep=1pt,minimum size=4pt] (n2) at (300: \R)  {1,1};
		\node[inner sep=1pt,minimum size=4pt] (n3) at (360: \R)  {3,3,4};
		
		\foreach \from/\to in {n6/n4,n5/n1,n2/n1,n5/n6,n5/n3,n6/n3}
		\draw[] (\from) -- (\to);
		\end{tikzpicture}
	}
	\hspace*{\fill}
	
	\centering
	\hspace*{\fill}
	\subfigure[Label hashing in $\mathcal{G}_1$.]{
		\begin{tikzpicture}
		[scale=1,every node/.style={draw, circle, minimum size=1.5em,inner sep=1}]
		\node (n6) at (60: \R) {8};
		\node (n4) at (120: \R)  {10};
		\node (n5) at (180: \R)  {11};
		\node (n1) at (240: \R) {7};
		\node (n2) at (300: \R)  {7};
		\node (n3) at (360: \R)  {9};
		
		\foreach \from/\to in {n4/n5,n5/n1,n2/n5,n3/n4,n5/n3,n6/n3}
		\draw[] (\from) -- (\to);
		\end{tikzpicture}
	}
	\hfill
	\centering
	\subfigure[Label hashing in $\mathcal{G}_2$.]{
		\begin{tikzpicture}
		[scale=1,every node/.style={draw, circle, minimum size=1.5em,inner sep=1}]
		\node (n6) at (60: \R) {9};
		\node (n4) at (120: \R)  {8};
		\node (n5) at (180: \R)  {12};
		\node (n1) at (240: \R) {6};
		\node (n2) at (300: \R)  {5};
		\node (n3) at (360: \R)  {10};
		
		\foreach \from/\to in {n6/n4,n5/n1,n2/n1,n5/n6,n5/n3,n6/n3}
		\draw[] (\from) -- (\to);
		\end{tikzpicture}
	}
	\hspace*{\fill}
	\caption{Illustration of one iteration of the WL test of graph isomorphism algorithm for graphs. For subfigures (a) and (b), $\Sigma=\{1,2,3,4\}$. For subfigures (e) and (f), $\Sigma=\{5,6,7,8,9,10,11,12\}$}
	\label{fig:wlcoloring}
\end{figure}

Let $\mathsf{P}$ denote the multiset of all shortest-paths in graph $\mathcal{G}$. For each shortest-path $P=(s,v_1,v_2,\ldots,t) \in\mathsf{P}$ where $s$ denotes the source vertex and $t$ denotes the sink vertex, we use a triplet $\left( l(s),l(t),len(P)\right) $ to denote it, where $len(P)$ is the length of the shortest-path $P$. For example, in Figure~\ref{fig:wlcoloring}(b), the triplet for the shortest-path between the two vertices with labels 2 and 4 respectively is (2,4,2). Then, for graph $\mathcal{G}$, its feature map is defined as follows:
\begin{equation}
\label{equ:shortest-path}
\phi(\mathcal{G})=\left[ \psi(\mathcal{G},S_1),\psi(\mathcal{G},S_2),\ldots,\psi(\mathcal{G},S_m)\right]
\end{equation}
where $m$ denotes the number of unique triplets in $\mathsf{P}$, and $\psi(\mathcal{G},S_i) (1\leq i\leq m)$ denotes the number of a unique triplet $S_i$ occurring in graph $\mathcal{G}$.

The Weisfeiler-Lehman test of graph isomorphism~\cite{weisfeiler1968reduction} belongs to the family of color refinement algorithms that iteratively update vertex colors (labels) until reaching the fixed number of iterations, or the vertex label sets of two graphs differ. In each iteration, the Weisfeiler-Lehman test of graph isomorphism algorithm augments vertex labels by first concatenating their neighbors' labels and then hashing the augmented labels into new labels. The hashed labels correspond to subtree patterns.

For example, in Figure~\ref{fig:wlcoloring}(b), a subtree pattern of height one rooted at the vertex with label 4 can be denoted as a string of concatenated labels of vertices ``4,1,3,3'' which is augmented as ``12' by the Weisfeiler-Lehman test of graph isomorphism. Let $\mathsf{T}^{(h)}$ denote the multiset of all subtree patterns of height $h$ in graph $\mathcal{G}$, then the feature map of $\mathcal{G}$ is defined as follows:
\begin{equation}
\label{equ:subtree}
\phi(\mathcal{G}^{(h)})=\left[ \psi(\mathcal{G}^{(h)},T_1^{(h)}),\psi(\mathcal{G}^{(h)},T_2^{(h)}),\ldots,\psi(\mathcal{G}^{(h)},T_m^{(h)})\right]
\end{equation}
where $\mathcal{G}^{(0)}$ is the original graph $\mathcal{G}$ and $\mathcal{G}^{(h)}$ is the augmented graph at the $h$-th iteration of the Weisfeiler-Lehman test of graph isomorphism. We call graphs $\mathcal{G}^{(0)},\mathcal{G}^{(1)},\ldots,\mathcal{G}^{(h)}$ a sequence of Weisfeiler-Lehman graphs. $m$ denotes the number of unique subtree patterns in $\mathsf{T}^{(h)}$, and $\psi(\mathcal{G}^{(h)},T_i^{(h)}) (1\leq i\leq m)$ denotes the number of a unique subtree pattern $T_i^{(h)}$ occurring in graph $\mathcal{G}^{(h)}$.

The feature map of WL is the concatenation of the feature maps at all the iterations:
 \begin{equation}
 \label{equ:wl}
 \phi(\mathcal{G})=\left[ \phi(\mathcal{G}^{(0)}),\phi(\mathcal{G}^{(1)}),\ldots,\phi(\mathcal{G}^{(h)})\right]
 \end{equation}

\section{Deep Graph Feature Maps}
In this section, we develop a new convolutional neural network (CNN)  model for learning deep graph feature maps, which is called \textsc{DeepMap}. The extension of CNNs from images whose pixels are spatially ordered to graphs of arbitrary size and shape is challenging. We first align vertices across graphs. Then, we build the receptive field for each vertex.

\comment{
\subsection{A Multiscale Graph Feature Map and Its Corresponding Graph Kernel}
From the analysis in Section \ref{pre}, we can see that the feature maps of the three popular graph kernels belong to our Definition \ref{def:fmg}. However, the three graph kernels cannot compare graphs at multiple levels of granularities. In the real world, many graphs such as molecules have structures at multiple levels of granularities. Graph kernels should not only capture the overall shape of graphs (whether they are more like a chain, a ring, a chain that branches, etc.), but also small structures of graphs such as chemical bonds and functional groups. For example, WL can only capture the graph similarity at coarse granularities, because subtrees can only consider neighborhood structures of vertices. SP can only capture the graph similarity at fine granularities, because shortest-paths do not consider neighborhood structures. To capture graph similarity at multiple levels of granularities, we propose a new feature map  and develop a new graph kernel called the subtree pattern graph kernel. The new feature map is based on an extension of the subtree patterns used in WL. WL considers only the shape of the whole subtree. Instead, we consider not only the whole shape of the subtree but also its child subtrees.

We first define the subtree pattern as follows:
\begin{definition}[Subtree Pattern]
	Given an undirected labeled graph $\mathcal{G}=(\mathcal{V},\mathcal{E}, l)$, we build a truncated tree $T$ of depth one rooted at each vertex $v \in\mathcal{V}$. The vertex $v$ is called root. For each subtree of $T$, we assign it a signature which can be denoted as a string of the concatenated labels of all its vertices, e.g., ``$l(v),l(v_1),\cdots,l(v_n)$''. Each signature is called a subtree pattern of root $v$.
\end{definition}

Figure \ref{fig:subtree}(d) shows a truncated tree of depth one rooted at the vertex with label 4. All the subtrees of this truncated tree are also shown in Figure \ref{fig:subtree}. All the subtree patterns are as follows: $T_1$=``4'', $T_2$=``4,1'', $T_3$=``4,1,3'', $T_4$=``4,1,3,3'', $T_5$=``4,3'' and $T_6$=``4,3,3''.  Note that the child vertices of the root are arranged according to their label values. Then, the subtree pattern feature map for a graph $\mathcal{G}$ is $\phi(\mathcal{G})=\left[ \psi(\mathcal{G},T_1),\psi(\mathcal{G},T_2),\ldots,\psi(\mathcal{G},T_m)\right]$, where $m$ is the number of the unique subtree patterns. 

\begin{figure}[htb]
	\hspace*{\fill}
	\centering
	\subfigure[$T_1$]{
		\begin{tikzpicture}
		[scale=1,level distance=17.32mm,
		every node/.style={draw, circle, minimum size=1.5em,inner sep=1},
		level 1/.style={sibling distance=10mm},
		level 2/.style={sibling distance=8mm}]
		\node {4};
		\end{tikzpicture}
	}
	\hfill
	\centering
	\subfigure[$T_2$]{
		\begin{tikzpicture}
		[scale=1,level distance=17.32mm,
		every node/.style={draw, circle, minimum size=1.5em,inner sep=1},
		level 1/.style={sibling distance=10mm},
		level 2/.style={sibling distance=8mm}]
		\node {4}
		child {node {1}
			child[missing]
		};
		\end{tikzpicture}
	}
    \hfill
	\centering
	\subfigure[$T_3$]{
		\begin{tikzpicture}
		[scale=1,level distance=17.32mm,
		every node/.style={draw, circle, minimum size=1.5em,inner sep=1},
		level 1/.style={sibling distance=10mm},
		level 2/.style={sibling distance=8mm}]
		\node {4}
		child {node {1}
			child[missing]
		}
		child{node{3}
			child[missing]
		};
		\end{tikzpicture}
	}
    \hfill
    \centering
    \subfigure[$T_4$]{
    	\begin{tikzpicture}
    	[scale=1,level distance=17.32mm,
    	every node/.style={draw, circle, minimum size=1.5em,inner sep=0pt},
    	level 1/.style={sibling distance=10mm},
    	level 2/.style={sibling distance=8mm}]
    	\node {4}
    	child {node {1}
    		child[missing]
    	}
    	child {node {3}
    		child[missing]
    	}
    	child {node {3}
    		child[missing]
    	};
    	\end{tikzpicture}
    }
    \hfill
	\centering
	\subfigure[$T_5$]{
		\begin{tikzpicture}
		[scale=1,level distance=17.32mm,
		every node/.style={draw, circle, minimum size=1.5em,inner sep=1},
		level 1/.style={sibling distance=10mm},
		level 2/.style={sibling distance=80mm}]
		\node {4}
		child {node {3}
			child[missing]
		};
		\end{tikzpicture}
	}
    \hfill
	\subfigure[$T_6$]{
		\begin{tikzpicture}
		[scale=1,level distance=17.32mm,
		every node/.style={draw, circle, minimum size=1.5em,inner sep=1},
		level 1/.style={sibling distance=10mm},
		level 2/.style={sibling distance=8mm}]
		\node {4}
		child {node {3}
			child[missing]
		}
		child{node{3}
			child[missing]
		};
		\end{tikzpicture}
	}
	\hspace*{\fill}
	\caption{All the subtree patterns rooted at the vertex with label 4.}
	\label{fig:subtree}
\end{figure}

However, this feature map does not consider the structural identity of each vertex. Structural identity is used to define the class of vertices by considering the graph structure. In graphs, vertices are often associated with some functions that determine their roles in the graph. For example, each of the proteins in a protein-protein interaction (PPI) network has a specific function, such as enzyme, antibody, messenger, transport/storage, and structural component. Although such functions may also depend on the vertex and edge attributes, in this paper, we only consider their relations to the graph structure.

To incorporate the structural identity of each vertex into the subtree pattern feature map, we use the Weisfeiler-Lehman test of graph isomorphism to generate a sequence of Weisfeiler-Lehman graphs. Then, for each graph in the sequence, we extract all the subtree patterns and compute our subtree pattern feature map. Note that the Weisfeiler-Lehman test of graph isomorphism first utilizes the truncated tree of depth one to update vertex colors. Then, the augmented label will be counted in the next iteration. Thus, in each iteration, we do not count the occurrences of the subtree pattern that corresponds to the truncated tree of depth one.

We first sort subtree patterns lexicographically, from low to high. Then, for each unique subtree pattern, we count its occurrences in the graph. The definition for our subtree pattern graph kernel is defined as follows:
\begin{equation}
\label{equ:ktp}
\mathcal{K}_{stp}(\mathcal{G}_1, \mathcal{G}_2) = \left\langle \phi(\mathcal{G}_1), \phi(\mathcal{G}_2)\right\rangle 
\end{equation}
where $\phi(\mathcal{G}_i)=\left[ \phi(\mathcal{G}_i^{(0)}),\phi(\mathcal{G}_i^{(1)}),\ldots,\phi(\mathcal{G}_i^{(h)})\right]$ ($i=1,2$) and $\phi(\mathcal{G}^{(j)})=\left[ \psi(\mathcal{G}^{(j)},T_1^{(j)}),\psi(\mathcal{G}^{(j)},T_2^{(j)}),\ldots,\psi(\mathcal{G}^{(j)},T_m^{(j)})\right]$ ($0\leq j\leq h$). $m$ is the number of the unique subtree patterns in the union of the two multisets $\mathsf{T}_1^{(j)}$ and $\mathsf{T}_2^{(j)}$ corresponding to $\mathcal{G}_1^{(j)}$ and $\mathcal{G}_2^{(j)}$, respectively.

\begin{theorem}
	The subtree pattern graph kernel $\mathcal{K}_{stp}$ is positive semi-definite.
\end{theorem}

\begin{proof}
    Inspired by earlier works on graph kernels, we can readily verify that for any vector $\mathbf{x}\in\mathbb{R}^n$, we have the following:
	\begin{equation}
	\label{equ:semidefinite}
	\begin{aligned}
	\mathbf{x}^\intercal \mathcal{K}_{stp}\mathbf{x}&=\sum_{i,j=1}^nx_ix_j\mathcal{K}_{stp}(\mathcal{G}_i, \mathcal{G}_j)\\
	&=\sum_{i,j=1}^nx_ix_j\left\langle \phi(\mathcal{G}_i), \phi(\mathcal{G}_j)\right\rangle\\
	&=\left\langle \sum_{i=1}^nx_i\phi(\mathcal{G}_i),\sum_{j=1}^nx_j\phi(\mathcal{G}_j)\right\rangle\\
	&=\left\| \sum_{i=1}^n\sum_{j=0}^hx_i\phi(\mathcal{G}_i^{(j)})\right\| \geq0
	\end{aligned}
	\end{equation}
\end{proof}
}

\subsection{CNNs on Graphs}
We define the vertex feature maps as follows:
\begin{definition}[Vertex Feature Maps]
	\label{def:fmv}
	Define a map $\psi: \{v_1,v_2,\ldots,v_{|\mathcal{V}|}\} \times \Sigma \rightarrow \mathbb{N}$ where $v_i\in \mathcal{V} (1\leq i\leq|\mathcal{V}|)$ such that $\psi(v, A)$ is the number of occurrences of the atomic substructure $A$ that contains $v$ in graph $\mathcal{G}$. Then the feature map of vertex $v$ is defined as follows:
	\begin{equation}
	\label{equ:vertex}
	\phi(v)=\left[ \psi(v,A_1),\psi(v,A_2),\ldots,\psi(v,A_m)\right]
	\end{equation}
	where $m$ is the number of unique atomic substructures and depends on graphs.
\end{definition}


From Definitions~\ref{def:fmg} and~\ref{def:fmv}, we can observe that the feature map of a graph equals to the sum of the feature maps of all the vertices in that graph. Note that this pooling-like feature map is permutation-invariant and size-invariant. In other words, the feature map is invariant to the ordering of vertices and the sizes of graphs.
\begin{equation}
\label{equ:add}
\phi(\mathcal{G})=\sum_{i=1}^{|\mathcal{V}|}\phi(v_i)
\end{equation}

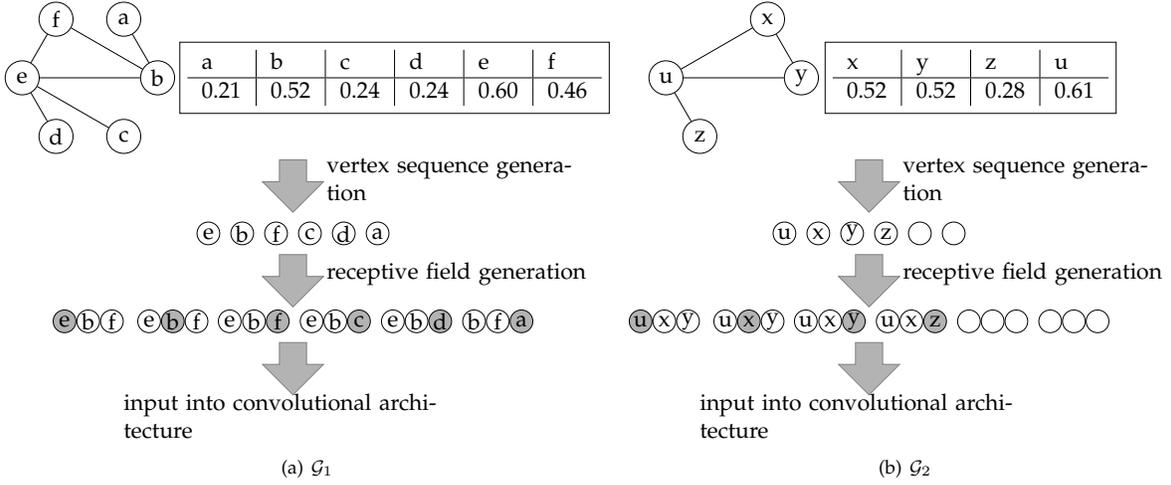
\begin{figure*}[htb]
	\centering
	\scalebox{0.9}{
	\hspace*{\fill}
	\subfigure[$\mathcal{G}_1$]{
		\begin{tikzpicture}
		\node (n6) at (-2.5, 0.866) [scale=1,draw, circle,minimum size=1.5em,inner sep=0pt] {a};
		\node (n4) at (-3.5,0.866) [scale=1,draw, circle,minimum size=1.5em,inner sep=0pt] {f};
		\node (n5) at (-4,0) [scale=1,draw, circle,minimum size=1.5em,inner sep=0pt] {e};
		\node (n1) at (-3.5,-0.866)[scale=1,draw, circle,minimum size=1.5em,inner sep=0pt] {d};
		\node (n2) at (-2.5,-0.866) [scale=1,draw, circle,minimum size=1.5em,inner sep=0pt] {c};
		\node (n3) at (-2,0) [scale=1,draw, circle,minimum size=1.5em,inner sep=0pt] {b};
		\foreach \from/\to in {n4/n5,n5/n1,n2/n5,n3/n4,n5/n3,n6/n3}
		\draw[] (\from) -- (\to);
		
		\node at (1.5,0) [shape=rectangle,draw] {
			\begin{tabular}{l|l|l|l|l|l}
			a &b &c &d &e &f \\\hline
			0.21&0.52&0.24&0.24&0.60 &0.46\\
			\end{tabular}
		};
		
		\node at (0,-1.5) [draw=gray,fill=gray!60,single arrow, rotate=-90] {\phantom{aa}};
		\node[text width=4cm] at (2.5,-1.5) {vertex sequence generation};
		
		\node at (-1.25,-2.3) [scale=1,draw, circle,minimum size=1em,inner sep=0pt] {e};
		\node at (-0.75,-2.3) [scale=1,draw, circle,minimum size=1em,inner sep=0pt] {b};
		\node at (-0.25,-2.3) [scale=1,draw, circle,minimum size=1em,inner sep=0pt] {f};
		\node at (0.25,-2.3)[scale=1,draw, circle,minimum size=1em,inner sep=0pt] {c};
		\node at (0.75,-2.3) [scale=1,draw, circle,minimum size=1em,inner sep=0pt] {d};
		\node at (1.25,-2.3) [scale=1,draw, circle,minimum size=1em,inner sep=0pt] {a};
		
		\node at (0,-2.9) [draw=gray,fill=gray!60,single arrow, rotate=-90] {\phantom{aa}};
		\node[text width=4cm] at (2.5,-2.9) {receptive field generation};
		\node at (-3.375,-3.6) [scale=1,draw, circle,fill=gray!60,minimum size=1em,inner sep=0pt] {e};
		\node at (-3.025,-3.6)[scale=1,draw, circle,minimum size=1em,inner sep=0pt]   {b};
		\node at (-2.675,-3.6)[scale=1,draw, circle,minimum size=1em,inner sep=0pt]   {f};
		\node at (-2.125,-3.6)[scale=1,draw, circle,minimum size=1em,inner sep=0pt]  {e};
		\node at (-1.775,-3.6) [scale=1,draw, circle,fill=gray!60,minimum size=1em,inner sep=0pt]  {b};
		\node at (-1.425,-3.6) [scale=1,draw, circle,minimum size=1em,inner sep=0pt]  {f};
		\node at (-0.935,-3.6)[scale=1,draw, circle,minimum size=1em,inner sep=0pt]  {e};
		\node at (-0.575,-3.6)[scale=1,draw, circle,minimum size=1em,inner sep=0pt]   {b};
		\node at (-0.225,-3.6) [scale=1,draw, circle,fill=gray!60,minimum size=1em,inner sep=0pt]  {f};
		\node at (0.275,-3.6)[scale=1,draw, circle,minimum size=1em,inner sep=0pt]  {e};
		\node at (0.625,-3.6)[scale=1,draw, circle,minimum size=1em,inner sep=0pt]   {b};
		\node at (0.975,-3.6) [scale=1,draw, circle,fill=gray!60,minimum size=1em,inner sep=0pt]  {c};
		\node at (1.475,-3.6)[scale=1,draw, circle,minimum size=1em,inner sep=0pt]  {e};
		\node at (1.825,-3.6) [scale=1,draw, circle,minimum size=1em,inner sep=0pt]  {b};
		\node at (2.175,-3.6) [scale=1,draw, circle,fill=gray!60,minimum size=1em,inner sep=0pt] {d};
		\node at (2.675,-3.6)[scale=1,draw, circle,minimum size=1em,inner sep=0pt]  {b};
		\node at (3.025,-3.6) [scale=1,draw, circle,minimum size=1em,inner sep=0pt]  {f};
		\node at (3.375,-3.6) [scale=1,draw, circle,fill=gray!60,minimum size=1em,inner sep=0pt]  {a};
		
		\node at (0,-4.2) [draw=gray,fill=gray!60,single arrow, rotate=-90] {\phantom{aa}};
		\node[text width=5cm] at (0,-5) {input into convolutional architecture};
		\end{tikzpicture}
	}
	\hfill
	\centering
	\subfigure[$\mathcal{G}_2$]{
		\begin{tikzpicture}
		\node (n6) at (-1.5, 0.866)[scale=1,draw, circle,minimum size=1.5em,inner sep=0pt] {x};
		\node (n5) at (-3,0) [scale=1,draw, circle,minimum size=1.5em,inner sep=0pt] {u};
		\node (n1) at (-2.5,-0.866) [scale=1,draw, circle,minimum size=1.5em,inner sep=0pt]{z};
		\node (n3) at (-1,0)[scale=1,draw, circle,minimum size=1.5em,inner sep=0pt]  {y};
		\foreach \from/\to in {n5/n1,n5/n6,n5/n3,n6/n3}
		\draw[] (\from) -- (\to);
		
		\node at (1.5,0) [shape=rectangle,draw] {
			\begin{tabular}{l|l|l|l}
			x &y &z &u \\\hline
			0.52&0.52&0.28&0.61\\
			\end{tabular}
		};
		
		\node at (0,-1.5) [draw=gray,fill=gray!60,single arrow, rotate=-90] {\phantom{aa}};
		\node[text width=4cm] at (2.5,-1.5) {vertex sequence generation};
		
		\node at (-1.25,-2.3) [scale=1,draw, circle,minimum size=1em,inner sep=0pt] {u};
		\node at (-0.75,-2.3) [scale=1,draw, circle,minimum size=1em,inner sep=0pt] {x};
		\node at (-0.25,-2.3) [scale=1,draw, circle,minimum size=1em,inner sep=0pt] {y};
		\node at (0.25,-2.3)[scale=1,draw, circle,minimum size=1em,inner sep=0pt] {z};
		\node at (0.75,-2.3)[scale=1,draw, circle,minimum size=1em,inner sep=0pt]  {};
		\node at (1.25,-2.3) [scale=1,draw, circle,minimum size=1em,inner sep=0pt] {};
		
		\node at (0,-2.9) [draw=gray,fill=gray!60,single arrow, rotate=-90] {\phantom{aa}};
		\node[text width=4cm] at (2.5,-2.9) {receptive field generation};
		
		\node at (-3.375,-3.6) [scale=1,draw, circle,fill=gray!60,minimum size=1em,inner sep=0pt] {u};
		\node at (-3.025,-3.6)[scale=1,draw, circle,minimum size=1em,inner sep=0pt] {x};
		\node at (-2.675,-3.6) [scale=1,draw, circle,minimum size=1em,inner sep=0pt] {y};
		\node at (-2.125,-3.6) [scale=1,draw, circle,minimum size=1em,inner sep=0pt] {u};
		\node at (-1.775,-3.6) [scale=1,draw, circle,fill=gray!60,minimum size=1em,inner sep=0pt]  {x};
		\node at (-1.425,-3.6) [scale=1,draw, circle,minimum size=1em,inner sep=0pt] {y};
		\node at (-0.935,-3.6)[scale=1,draw, circle,minimum size=1em,inner sep=0pt] {u};
		\node at (-0.575,-3.6) [scale=1,draw, circle,minimum size=1em,inner sep=0pt] {x};
		\node at (-0.225,-3.6) [scale=1,draw, circle,fill=gray!60,minimum size=1em,inner sep=0pt]  {y};
		\node at (0.275,-3.6)[scale=1,draw, circle,minimum size=1em,inner sep=0pt] {u};
		\node at (0.625,-3.6) [scale=1,draw, circle,minimum size=1em,inner sep=0pt] {x};
		\node at (0.975,-3.6) [scale=1,draw, circle,fill=gray!60,minimum size=1em,inner sep=0pt]  {z};
		\node at (1.475,-3.6)[scale=1,draw, circle,minimum size=1em,inner sep=0pt] {};
		\node at (1.825,-3.6) [scale=1,draw, circle,minimum size=1em,inner sep=0pt] {};
		\node at (2.175,-3.6)[scale=1,draw, circle,minimum size=1em,inner sep=0pt] {};
		\node at (2.675,-3.6)[scale=1,draw, circle,minimum size=1em,inner sep=0pt] {};
		\node at (3.025,-3.6) [scale=1,draw, circle,minimum size=1em,inner sep=0pt] {};
		\node at (3.375,-3.6) [scale=1,draw, circle,minimum size=1em,inner sep=0pt] {};
		
		\node at (0,-4.2) [draw=gray,fill=gray!60,single arrow, rotate=-90] {\phantom{aa}};
		\node[text width=5cm] at (0,-5) {input into convolutional architecture};
		\end{tikzpicture}
	}
	\hspace*{\fill}
    }
	
	\caption{Illustration of the generation of vertex sequences and the corresponding receptive fields for undirected graphs $\mathcal{G}_1$ and $\mathcal{G}_2$. The two tables beside graphs $\mathcal{G}_1$ and $\mathcal{G}_2$ contain the eigenvector centrality value of each vertex.}
	\label{fig:generation}
\end{figure*}

\begin{figure*}[!htp]
	\centering
	\includegraphics[width=0.8\textwidth]{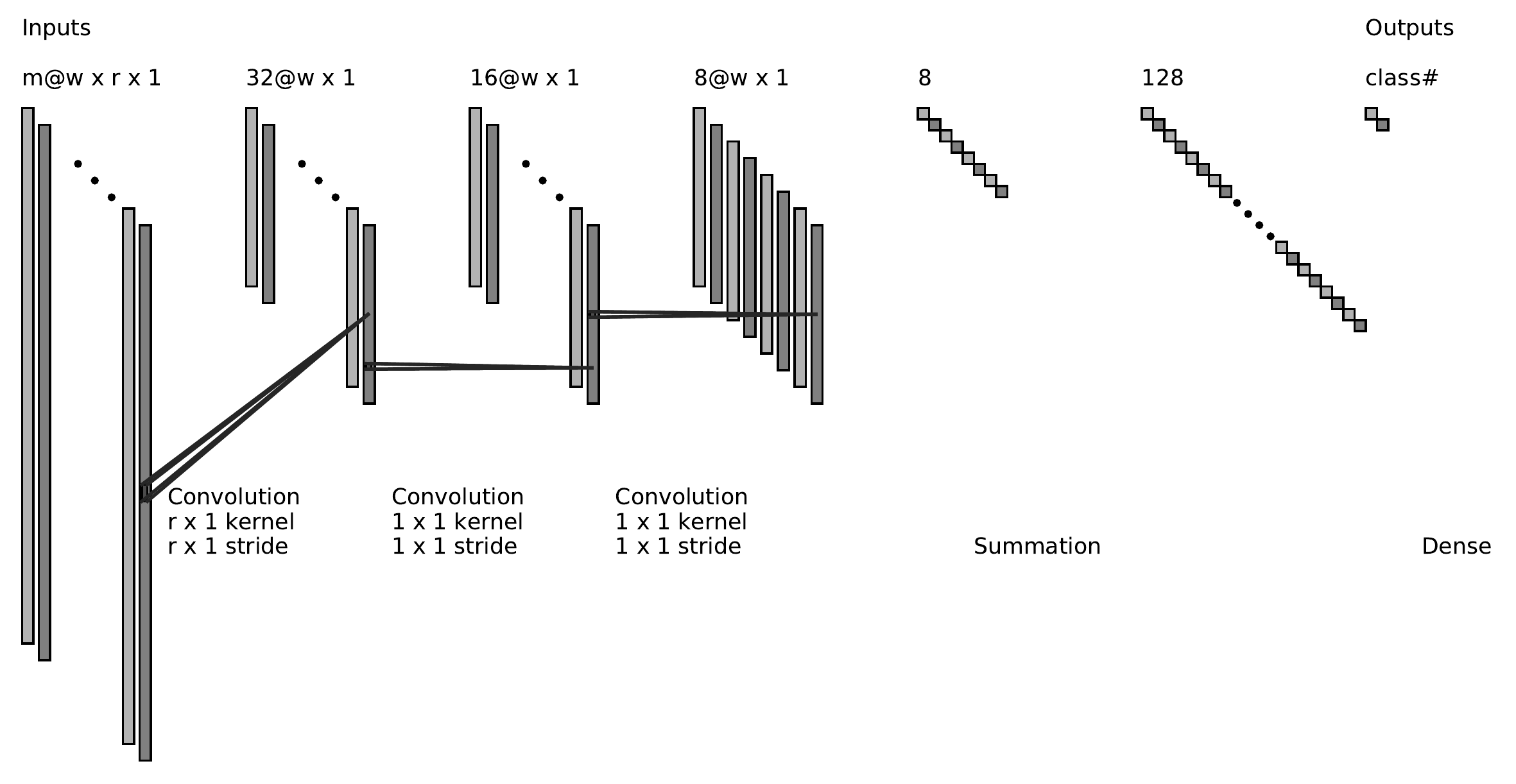}
	\caption{Convolutional architecture. In the first layer, $m$ is the dimension of the vertex feature map, $w$ is the maximum number of vertices in a set of graphs, and $r$ is the size of the receptive field.}
	\label{fig:cnn}
\end{figure*}

From the definitions of vertex feature maps (Equation~\ref{equ:vertex}) and graph feature maps (Equation~\ref{equ:graph}), we can see that several important issues are not taken into account to compute graph feature maps. As described in Section~\ref{intro}, they are: (1) Substructures are not independent and thus it leads to high-dimensional feature space. (2) The complex high-order interactions in the neighborhood of a vertex are not considered. To mitigate these two issues, in this work, we develop a CNN architecture on the vertex feature maps. The learned deep graph representation is of low-dimension. Furthermore, convolution operation in CNN can capture the complex high-order interactions in the neighborhood of a vertex. One main challenge in developing CNNs for graphs of arbitrary size and shape is that unlike images whose pixels are spatially ordered, vertices in graphs do not have a spatial or temporal order. Vertices across different graphs are difficult to align, and thus the receptive fields of CNNs cannot be directly applied on vertices in graphs.

An image can be considered as a rectangle grid graph whose vertices represent pixels. A CNN of a stride length one on an image can be considered as traversing a sequence of pixels (vertices), from left to right and top to bottom. As indicated above, pixels are spatially ordered and they are aligned across images. Thus, the order of pixels in the sequence that is traversed by a CNN is unique. To make CNNs applicable to graphs, we first need to generate a vertex sequence for each graph such that the sequences are aligned across graphs. In this work, we use eigenvector centrality~\cite{bonacich1987power} to measure the importance of a vertex. A vertex has high eigenvector centrality value if it is linked to by other vertices that also have high eigenvector centrality values, without implying that this vertex is highly linked. We generate a vertex sequence in each graph by sorting their eigenvector centrality values from high to low. Since graphs are of arbitrary size, we use the size $w$ of the graph that has the largest number of vertices as the length of the sequence. In this case, for sequences whose lengths are less than $w$, we concatenate them with dummy vertices to make their lengths equal to $w$. The dummy vertices' feature maps are set to zero vectors so that they do not contribute to the convolution.

After generating a vertex sequence for each graph, we need to determine the receptive field for each vertex in the sequence. Assume that the size of the receptive field is $r$. We perform a breadth-first search (BFS) on the original graph for constructing the receptive field. If the number of the one-hop neighbors of a vertex is greater than or equal to $r-1$, we select the top $r-1$ largest one-hop neighbors with respect to their eigenvector centrality values. If the number of the one-hop neighbors of a vertex is less than $r-1$, we first select all the one-hop neighbors and then select vertices from the two-hop neighbors, the three-hop neighbors, and so on, until the receptive field has exact $r$ vertices. If the size of a graph is less than $r$, we use dummy vertices for padding purposes. Note that the vertices in the receptive field are also sorted in descending order according to their eigenvector centrality values. 

We use Figure~\ref{fig:generation} to demonstrate the generation procedure for vertex sequences and their corresponding receptive fields. In the first row of Figure~\ref{fig:generation}, the tables demonstrate the eigenvector centrality value of each vertex. In the second row of Figure~\ref{fig:generation}, the vertex sequence is generated by sorting the eigenvector centrality values of vertices in descending order. Since the size of graph $\mathcal{G}_2$ is less than that of graph $\mathcal{G}_1$, we concatenate two dummy vertices (indicated by two blank vertices) in the generated sequence. In the third row of Figure~\ref{fig:generation}, for each vertex (indicated in gray) in the vertex sequence, we use BFS to generate its receptive field. Here, the size of the receptive field is three. Finally, we use a convolutional architecture to learn deep graph feature maps from the feature maps of the vertices.

Figure~\ref{fig:cnn} demonstrates our convolutional architecture. The architecture has three one-dimensional convolution layers which have rectified linear units (ReLU). Three one-dimensional convolution layers are used to aggregate the feature maps of each vertex with those of its neighbors. After the last convolution layer, we use a summation layer to add the feature map of every vertex in a graph together and the resulting feature map is the deep graph feature map. This summation layer just functions Equation~\ref{equ:add}. After the summation layer, we use a dense (fully-connected) layer with rectified linear units (ReLU), followed by a dropout layer and a softmax layer, for graph classification.


%

 \begin{theorem}
	If two graphs $\mathcal{G}_1$ and $\mathcal{G}_2$ are isomorphic, their deep graph feature maps after the summation layer are the same. 
 \end{theorem}

\begin{proof}
	If $\mathcal{G}_1 \simeq \mathcal{G}_2$, we have the following:
	\begin{itemize}
	    \item $|\mathcal{V}_1|=|\mathcal{V}_2|$ and $|\mathcal{E}_1|=|\mathcal{E}_2|$.
	    \item $\mathcal{G}_1$ and $\mathcal{G}_2$ have the same degree sequence.
		\item Vertex $v_1$ from $\mathcal{G}_1$ and vertex $\varphi(v_1)$ from $\mathcal{G}_2$ have the same feature map $\phi(v_1) = \phi(\varphi(v_1))$.
		\item Vertex sequence $S_1 = \left( v_1, v_2,\cdots, v_{|\mathcal{V}_1|}\right) $ generated from $\mathcal{G}_1$ is identical to vertex sequence $S_2 = \left(\varphi(v_1), \varphi(v_2),\cdots, \varphi(v_{|\mathcal{V}_2|})\right) $ generated from $\mathcal{G}_2$.
		 \item The receptive field for each vertex in sequence $S_1$ is the same as that of the corresponding vertex in sequence $S_2$.	 
	\end{itemize}
    Thus, $\mathcal{G}_1$ and $\mathcal{G}_2$ have the same deep graph feature maps after the summation layer.
\end{proof}

Note that if we use the sampling technique to sample graphlets around two corresponding vertices from two isomorphic graphs, the vertex feature maps may not be the same. Thus, the deep graph feature maps may not be the same.

\subsection{Algorithm}
The pseudo-code for \textsc{DeepMap} is given in Algorithm 1. Lines 1--7 compute the feature map for each vertex in each graph. The feature map could be graphlet feature map, shortest-path feature map, or subtree feature map. The user can choose one kind of them. If using random sampling for graphlets, the time complexity to compute feature maps for all vertices in $n$ graphs is $\mathcal{O}(n\cdot w\cdot d^3)$~\cite{shervashidze2009efficient}, where $w$ is the largest number of vertices in a set of graphs, and $d$ is the maximum degree number. If using the Floyd–Warshall algorithm to find all pairs of shortest paths, the time complexity to compute feature maps for all vertices in $n$ graphs is $\mathcal{O}(n\cdot w^3)$~\cite{borgwardt2005shortest}. If using subtrees, the time complexity to compute feature maps for all vertices in $n$ graphs is $\mathcal{O}(n\cdot h\cdot e)$~\cite{shervashidze2011weisfeiler}, where $e$ is the largest number of edges and we assume $e>w$, and $h$ is the iteration of the Weisfeiler-Lehman test of graph isomorphism.

For each graph, line 11 generates the vertex sequence by sorting vertices in descending order with respect to their eigenvector centrality values. We use power iteration to compute eigenvector centrality for each graph. The time complexity for line 11 is bounded by $\mathcal{O}(e+w\cdot\log w)$, where $\mathcal{O}(e)$ is the time complexity of power iteration, and $\mathcal{O}(w\cdot\log w)$ is the time complexity for the fast sort. If the size of a graph is less than the designated length $w$ of the vertex sequence, line 13 appends its vertex sequence with dummy vertices. For each vertex $v$ in the sequence, lines 15--19 construct its receptive field and append the corresponding vertex feature maps to $\matrixSym{\Phi}'$. Line 17 uses the breadth-first search (BFS) starting from $v$ on the original graph to find the top $r-1$ largest neighbors w.r.t. their eigenvector centrality values, and sort them in descending order. We assume the number of edges is greater than the number of vertices. Thus, the time complexity for BFS is bounded by $\mathcal{O}(e)$.  Thus, the time complexity for lines 10--20 is $\mathcal{O}\left( n\cdot(e+w\cdot\log w + w\cdot e)\right)$. Finally, we input $\matrixSym{\Phi}'$ and the graph labels $\mathcal{Y}$ into CNNs for graph classification. 
 
The time complexity of two-dimensional CNNs is $\mathcal{O}\left( \sum_{l=1}^cn_{ic}\cdot s_l^2\cdot n_{f}\cdot m_{oc}^2\right)$~\cite{he2015convolutional}, where $l$ is the index of a convolutional layer, $c$ is the number of convolutional layers, $n_{ic}$ is the number of input channels, $s_l$ is the length of the receptive field, $n_{f}$ is the number of filters in the $l$-th layer, $m_{oc}$ is the size of output channels. In our one-dimensional CNNs, $c$ is set to three. In the first layer, $n_{ic}$ is the length $m$ of the vertex feature maps extracted by counting the substructures around the vertices, $s_l$ is $r$, $n_{f}$ is $w$, $m_{oc}$ is set to 32. In the second layer, $n_{ic}$ is 32, $s_l$ is set to one, $n_{f}$ is $w$, $m_{oc}$ is set to 16. In the third layer, $n_{ic}$ is 16, $s_l$ is also set to one, $n_{f}$ is $w$, $m_{oc}$ is set to 8. Thus, the time complexity of our one-dimensional CNNs is bounded by $\mathcal{O}(m\cdot r\cdot w)$. The dense layer has 128 units. The dropout layer has a dropout rate of 0.5. The worst-case (when using the shortest-path feature map) time complexity of \textsc{DeepMap} is $\mathcal{O}(n\cdot w^3 + n\cdot w \cdot e + m\cdot w\cdot r)$.

\begin{algorithm2e}
	\SetKwFunction{cnn}{CNNs}
	\KwIn{A set of  graphs $\{\mathcal{G}_1, \mathcal{G}_2, \cdots, \mathcal{G}_n\}$ and their corresponding labels $\mathcal{Y}=\{y_1,y_2,\cdots,y_n\}$, the size $r$ of the receptive field}
	\KwOut{Classification accuracy $acc$}
	$\matrixSym{\Phi}\leftarrow []$\tcc*[r]{feature maps initialization.}
	\ForEach{\upshape graph $\mathcal{G}_i (1\leq i\leq n)$ }{
		$\matrixSym{X}\leftarrow[]$\;
		\ForEach{\upshape vertex $v\in\mathcal{G}_i$ }{
		$\mathbf{x}\leftarrow\phi(v)$\;
		$\matrixSym{X}$.append($\mathbf{x}$)\;
	    }
        $\matrixSym{\Phi}$.append($\matrixSym{X}$)\;
	}
	$w\leftarrow\max_{\mathcal{G}_i}$ \upshape length($\mathcal{G}_i$) ($1\leq i\leq n$)\;
	$\matrixSym{\Phi}'\leftarrow[]$\;
	\ForEach{\upshape graph $\mathcal{G}_i (1\leq i\leq n)$ }{
		generate vertex sequence $S_i=\left( v_{\sigma_1},v_{\sigma_2},\cdots,v_{\sigma_{|\mathcal{V}_i|}}\right) $ by sorting vertices according to their eigenvector centrality values\tcc*[r]{$\sigma_1,\sigma_2,\cdots,\sigma_{|\mathcal{V}_i|}$ is a permutation of $1,2,\cdots,|\mathcal{V}_i|$.}
		\If{\upshape length($\mathcal{G}_i$)$<w$}{
			append $w-$\upshape length($\mathcal{G}_i$) dummy vertices to $S_i$\;
		}
	    $\matrixSym{X}'\leftarrow[]$\;
	    \ForEach{\upshape vertex $v$ \upshape in \upshape sequence $S_i$ }{
	    	\If{$v$ \upshape is \upshape not \upshape a \upshape dummy \upshape vertex}{
	    		append $\matrixSym{X}(v_{\sigma_1}),\matrixSym{X}(v_{\sigma_2}),\cdots,\matrixSym{X}(v),\cdots,\matrixSym{X}(v_{\sigma_{r-1}})$ to $\matrixSym{X}'$\tcc*[r]{use the breadth-first search (BFS) starting from $v$ on the original graph to find the top $r-1$ largest neighbors w.r.t. their eigenvector centrality values, and sort them in descending order.}
	    	}
    	    \Else{
    	        append number $r$ of zero vectors $\mathbf{0}$ (for dummy vertices) to $\matrixSym{X}'$\;
            }
	    }
    $\matrixSym{\Phi}'$.append($\matrixSym{X}'$)\;
	}
    $acc\leftarrow$\cnn{$\matrixSym{\Phi}'$,$\mathcal{Y}$}\tcc*[r]{10-fold cross-validation.}
	\Return{$acc$} \;
	\caption{\textsc{DeepMap}}
	\label{alg:deepmap}
\end{algorithm2e}

\section{Experimental Evaluation}\label{experiments}
In this section, we conduct experiments on the benchmark graph datasets to compare \textsc{DeepMap} with state-of-the-art graph kernels and GNNs.  \textsc{DeepMap} is built on three kinds of vertex feature maps: graphlet (\textsc{GK}~\cite{shervashidze2009efficient}), shortest-path (\textsc{SP}~\cite{borgwardt2005shortest}), and subtree patterns (\textsc{WL}~\cite{shervashidze2011weisfeiler}). The corresponding three versions of \textsc{DeepMap} are denoted as \textsc{DeepMap-GK}, \textsc{DeepMap-SP}, and \textsc{DeepMap-WL}, respectively.

\subsection{Experimental Setup}
We run all the experiments on a server with a 32-core Intel(R) Xeon(R) Silver 4110 CPU@2.10GHz, 128 GB memory, a quad-core GeForce RTX 2080 GPU, and Ubuntu 18.04.1 LTS operating system, Python version 3.6. \textsc{DeepMap} is implemented with the Tensorflow wrapper Keras. We make our code publicly available at Github\footnote{\url{https://github.com/yeweiysh/DeepMap}}.

We compare \textsc{DeepMap} with six state-of-the-art graph kernels, i.e.,  \textsc{GNTK}~\cite{du2019gntk}, \textsc{DGK}~\cite{yanardag2015deep}, \textsc{RetGK}~\cite{zhang2018retgk}, \textsc{GK}~\cite{shervashidze2009efficient}, \textsc{SP}~\cite{borgwardt2005shortest}, and \textsc{WL}~\cite{shervashidze2011weisfeiler}. We also compare \textsc{DeepMap} with four state-of-the-art GNNs, i.e., \textsc{GIN}~\cite{xu2018powerful}, \textsc{PatchySan}~\cite{niepert2016learning}, \textsc{DCNN}~\cite{atwood2016diffusion}, and \textsc{DGCNN}~\cite{zhang2018end}. We perform 10-fold cross-validation and report the average classification accuracies and standard deviations. 

For the comparison methods, we set their parameters according to their original papers. The graphlet size of \textsc{GK} is selected from $\{3, 4, 5\}$. The depth of the subtree used in \textsc{WL} is selected from $\{0, 1, 2, 3, 4, 5\}$. For \textsc{DeepMap}, we use a single network architecture for all the experiments. We use the RMSPROP optimizer with initial learning rate 0.01 and decay the learning rate by 0.5 if the number of epochs with no improvement in the loss reaches five. We select the number of batch size from $\{32, 256\}$. Following \textsc{GIN}~\cite{xu2018powerful}, for \textsc{DeepMap} and other GNNs, the number of epochs is set as the one that has the best cross-validation accuracy averaged over the ten folds. For graph kernels, we use a binary $C$-SVM~\cite{chang2011libsvm} as the classifier. The parameter $C$ for each fold is independently tuned from $\left\lbrace1,10,10^{2},10^{3}\right\rbrace $ using the training data from that fold.

\subsection{Datasets}
In order to test the effectiveness of \textsc{DeepMap}, we use benchmark datasets whose statistics are given in Table~\ref{tab:statistics_dataset}. For datasets without vertex labels, we use vertex degrees as their vertex labels.

\textbf{Synthetic dataset}. SYNTHIE~\cite{morris2016faster} contains 400 graphs and can be divided into four classes. They are generated from two Erd\H{o}s-R\'{e}nyi graphs with edge probability 0.2.

\textbf{Brain network dataset}. KKI~\cite{pan2017task} is a brain network constructed from the whole brain functional resonance image (fMRI) atlas. Each vertex corresponds to a region of interest (ROI), and each edge indicates correlations between two ROIs. KKI is constructed for the task of Attention Deficit Hyperactivity Disorder (ADHD) classification.

\textbf{Chemical compound datasets}. The chemical compound datasets BZR\_MD, COX2\_MD, and DHFR are from~\cite{sutherland2003spline}. Chemical compounds or molecules are represented by graphs. Edges represent the chemical bond type, i.e., single, double, triple or aromatic. Vertices represent atoms. Vertex labels represent atom types. BZR is a dataset of ligands for the benzodiazepine receptor. COX2 is a dataset of cyclooxygenase-2 inhibitors. DHFR is a dataset of 756 inhibitors of dihydrofolate reductase. BZR\_MD and COX2\_MD are derived from BZR and COX2, respectively, by removing explicit hydrogen atoms. The chemical compounds in the datasets BZR\_MD and COX2\_MD are represented as complete graphs. NCI1~\cite{wale2008comparison} is a balanced dataset of chemical compounds screened for the ability to suppress the growth of human non-small cell lung cancer.

\textbf{Molecular compound datasets}. The PTC~\cite{kriege2012subgraph} dataset consists of compounds labeled according to carcinogenicity on rodents divided into male mice (MM), male rats (MR), female mice (FM) and female rats (FR). ENZYMES is a dataset of protein tertiary structures from~\cite{borgwardt2005protein}, consisting of 600 enzymes from six Enzyme Commission top-level enzyme classes. The dataset PROTEINS is from~\cite{borgwardt2005protein}. Each protein is represented by a graph. Vertices represent secondary structure elements. Edges represent that two vertices are neighbors along the amino acid sequence or three-nearest neighbors to each other in space.

\textbf{Movie collaboration dataset}. IMDB-BINARY and IMDB-MULTI datasets are from~\cite{yanardag2015deep}. IMDB-BINARY contains movies of different actor/actress and genre information. For each collaboration graph, vertices represent actors/actresses. Edges denote that two actors/actresses appear in the same movie. The collaboration graphs are generated on Action and Romance genres. And for each actor/actress, a corresponding ego-network is derived and labeled with its genre. IMDB-MULTI is a multi-class version of IMDB-BINARY and includes a balanced set of ego-networks derived from Comedy, Romance, and Sci-Fi genres.

\textbf{Scientific collaboration dataset}. COLLAB~\cite{leskovec2005graphs} is derived from three public collaboration datasets, i.e., High Energy Physics, Condensed Matter Physics, and Astro Physics. Each graph represents an ego-network of a researcher from a research field. The label represents the research field (High Energy Physics, Condensed Matter Physics, and Astro Physics) of a researcher.

\begin{table}[!htb]
	\centering
	\caption{Statistics of the benchmark datasets used in the experiments. N / A means the dataset has no vertex labels.}
	\label{tab:statistics_dataset}
	\begin{tabular}{l|l|l|l|l|l}
		\toprule
		\multirow{2}{*}{Dataset}         &Size            & Class & Avg.    & Avg.   &Label\\ 
		                         &                      &\#             &Node\#           &Edge\#          &\# \\\hline
		SYNTHIE           &400                &4              &95.00                &172.93         &N / A\\
		KKI                    &83             &2               &26.96                &48.42        &190 \\
		BZR\_MD        &306                &2              &21.30                &225.06         &8            \\ 
		COX2\_MD     &303                &2             &26.28                &335.12         &7             \\
		DHFR               &467             &2               &42.43                &44.54         &9           \\ 
		NCI1                   &4110        &2                 &17.93                 &19.79        &37          \\
		PTC\_MM        &336           &2               &13.97                 &14.32         &20             \\
		PTC\_MR         &344          &2               &14.29                 &14.69        &18               \\
		PTC\_FM          &349           &2               &14.11                  &14.48        &18               \\
		PTC\_FR          &351            &2               &14.56                  &15.00        &19               \\
		ENZYMES        &600           &6               &32.63                &62.14         &3            \\
		PROTEINS         &1113        &2                 &39.06                 &72.82         &3           \\
		IMDB-BINARY  &1000        &2                &19.77                 &96.53        &N / A\\     
		IMDB-MULTI    &1500        &3                &13.00                 &65.94         &N / A\\ 
		COLLAB             &5000        &3                &74.49                &2457.78     &N / A            \\
		\bottomrule
	\end{tabular}
\end{table}

\subsection{Results}
In this section, we first evaluate \textsc{DeepMap} with varying the size $r$ of the receptive field, then compare \textsc{DeepMap} with baselines on representational power and classification accuracy. 

\subsubsection{Parameter Sensitivity}
\begin{figure}[!htp]
	\centering
	\includegraphics[width=0.35\textwidth]{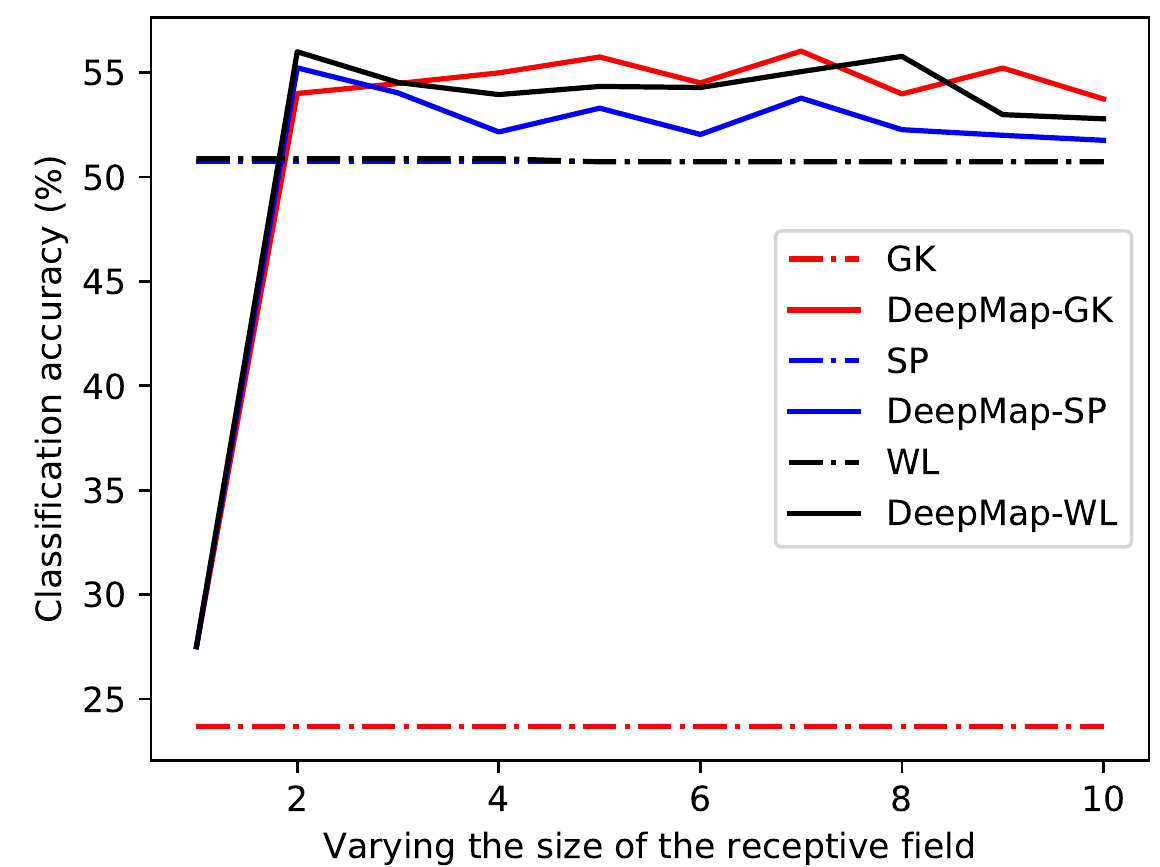}
	\caption{Parameter sensitivity studies for the deep map models and their corresponding graph kernels on the benchmark dataset SYNTHIE.}
	\label{fig:parameter}
\end{figure}

We test the parameter sensitivity of the deep map models and their corresponding graph kernels on the synthetic dataset SYNTHIE. The results are shown in Figure \ref{fig:parameter}. Because graph kernels do not have the parameter of the size $r$ of the receptive field, their classification accuracies do not change. When the size of the receptive field equals one, i.e., no neighborhood information is used in the deep map models, we can see that the deep map models perform poorly (classification accuracy is around 27\%). When the size of the receptive field exceeds two, all the three deep map models are superior to their corresponding graph kernels. The performance of \textsc{DeepMap-SP} decreases with the increasing size of the receptive field, which can be explained by the small world experiments \cite{milgram1967small}. The experiments are often associated with the phrase ``six degrees of separation'', which means every two vertices in a graph can be connected by a shortest-path of length at most six. When the size of the receptive field of a vertex exceed seven (including the vertex itself), extra neighbors deteriorate the discrimination power. We can also observe a similar tendency of \textsc{DeepMap-WL} because it is built on subtree patterns which are also constrained by the ``six degrees of separation''. For \textsc{DeepMap-GK}, its performance increases with the increasing size of the receptive field. \textsc{DeepMap-GK} is built on graphlets. For each vertex, we randomly sample 20 graphlets of size five. More neighbors provide more distinct information and thus improve the discrimination power.



\subsubsection{Representational Power}
\begin{figure}[!htp]
	\centering
	\includegraphics[width=0.35\textwidth]{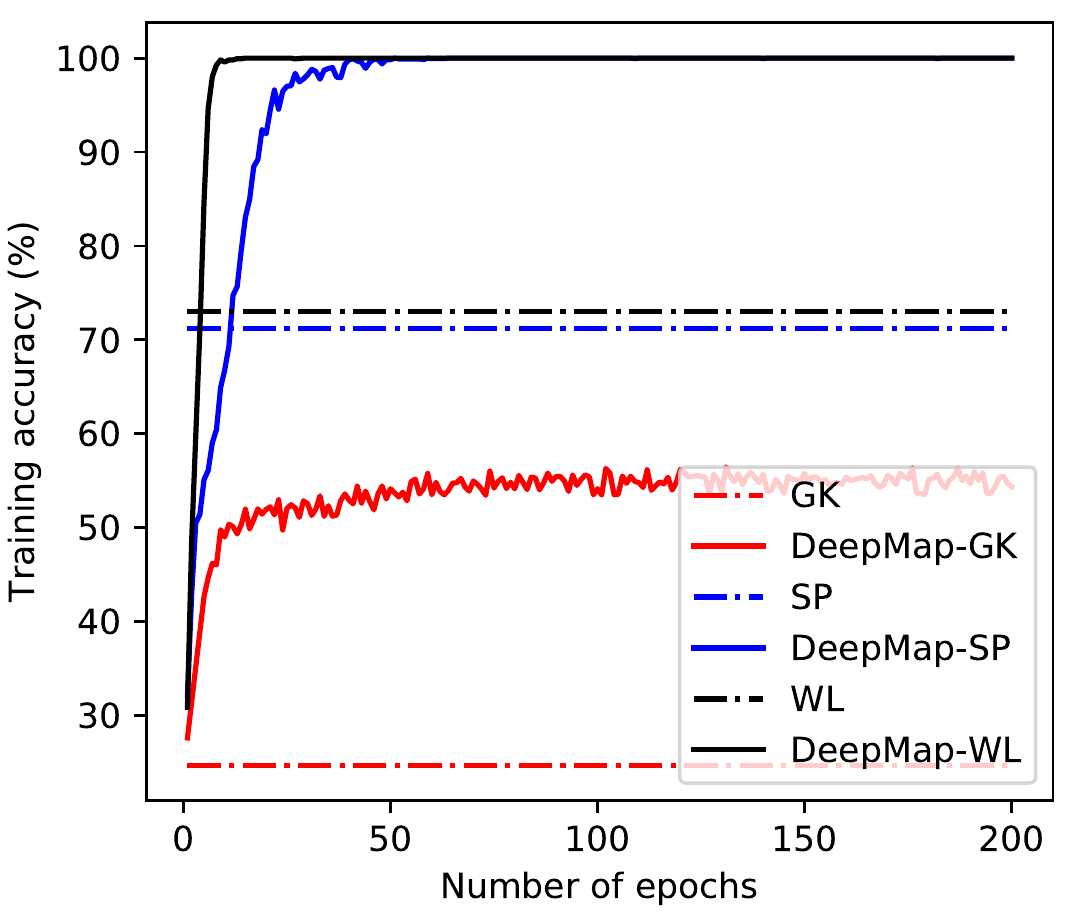}
	\caption{Representational power studies for the deep map models and their corresponding graph kernels on the benchmark dataset SYNTHIE.}
	\label{fig:rp1}
\end{figure}

\begin{figure}[!htp]
	\centering
	\includegraphics[width=0.35\textwidth]{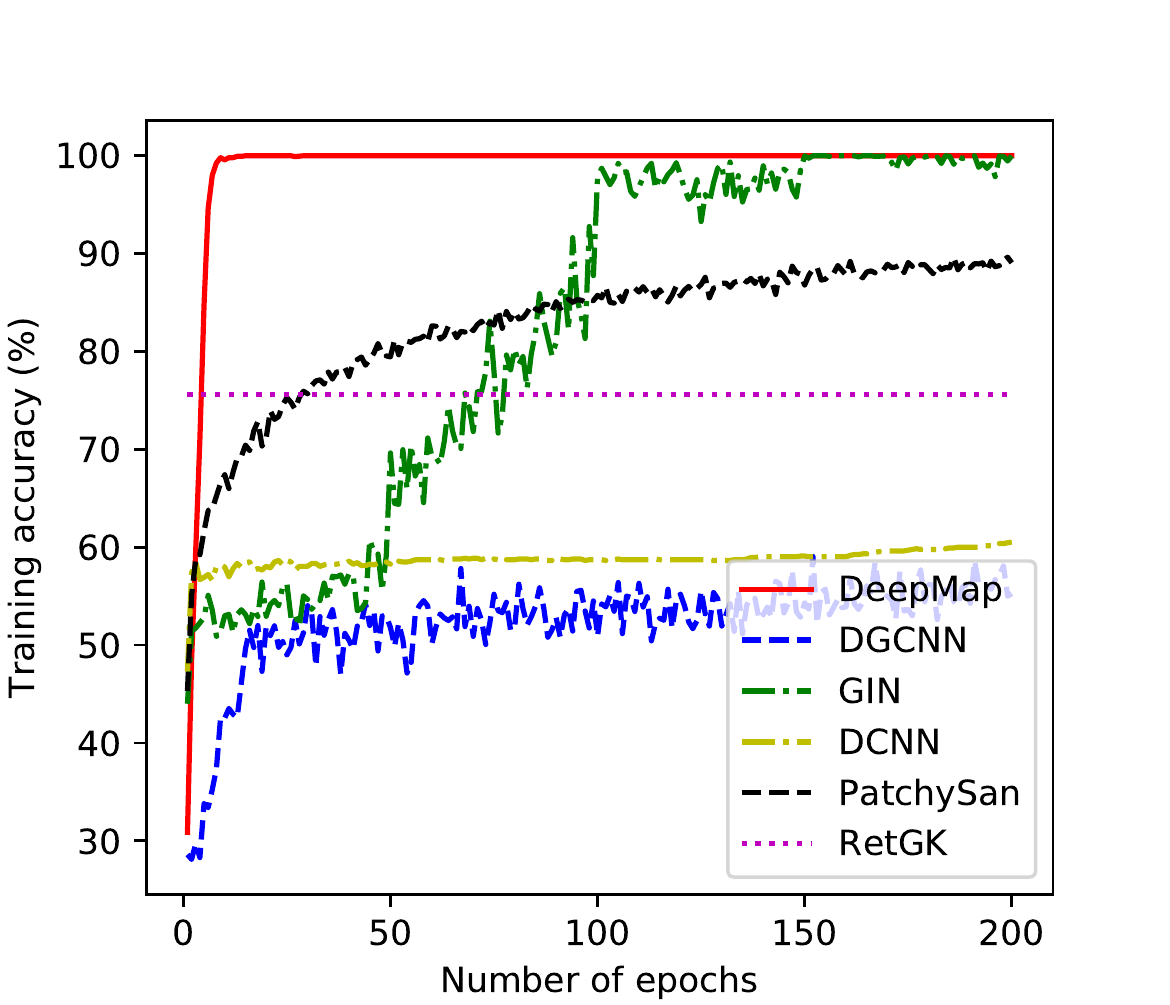}
	\caption{Representational power studies for \textsc{DeepMap} and other baselines on the benchmark dataset SYNTHIE. To reduce clutter, for graph kernels, we only show the result of the graph kernel that has the highest representational power.}
	\label{fig:rp2}
\end{figure}

We test the representational power of the deep map models and their corresponding graph kernels on the benchmark dataset SYNTHIE in Figure \ref{fig:rp1}. We use the average training accuracy over the ten folds to evaluate the representational power. We can see that the deep map models dramatically improve the representational power of their corresponding graph kernels. \textsc{DeepMap-WL} and \textsc{DeepMap-SP} converge faster than \textsc{DeepMap-GK}. We use the best results of the deep map models in Figure \ref{fig:rp1} as the final results for \textsc{DeepMap}. We compare the representational power of \textsc{DeepMap} with other baselines in Figure \ref{fig:rp2}. We can observe that \textsc{DeepMap} has better representational power and converges faster than other GNNs. \textsc{DeepMap} is superior to all the other baselines with a large margin.


%
%
%

%
%
%

 \begin{table*}[!htb]
	\centering
	\caption{Comparison of classification accuracy ($\pm$ standard deviation) of the deep map models to their corresponding graph kernels on the benchmark datasets.}
	\label{tab:classification1}
	\begin{tabular}{l|l|l||l|l||l|l}
		\toprule
		Dataset              &\textsc{GK}       &\textsc{DeepMap-GK}          &\textsc{SP}          &\textsc{DeepMap-SP}     &\textsc{WL}            &\textsc{DeepMap-WL} \\ \hline
		SYNTHIE             &23.68$\pm$2.11     &\textbf{54.48$\pm$4.34}    &50.73$\pm$1.74   &\textbf{54.03$\pm$2.38}     &50.88$\pm$1.04        &\textbf{54.53$\pm$6.16}   \\
		KKI                   &51.88$\pm$3.19     &\textbf{56.77$\pm$9.69}     &50.13$\pm$3.46   &\textbf{62.92$\pm$7.94}    &50.38$\pm$2.77        &\textbf{61.65$\pm$15.0}     \\
		BZR\_MD           &49.27$\pm$2.15     &\textbf{63.11$\pm$10.0}    &68.60$\pm$1.94   &\textbf{73.55$\pm$5.76}     &59.67$\pm$1.47       &\textbf{71.56$\pm$6.66}    \\ 
		COX2\_MD        &48.17$\pm$1.88     &\textbf{52.44$\pm$7.36}    &65.70$\pm$1.66    &\textbf{72.28$\pm$9.37}     &56.30$\pm$1.55         &\textbf{69.66$\pm$7.32}     \\ 
		DHFR                &61.01$\pm$0.23     &\textbf{61.64$\pm$2.07}    &77.80$\pm$0.98   &\textbf{81.35$\pm$4.08}     &82.39$\pm$0.90         &\textbf{85.17$\pm$2.19}     \\ 
		NCI1                 &62.11$\pm$0.19        &\textbf{63.26$\pm$2.04}    &73.12$\pm$0.29    &\textbf{79.90$\pm$1.78}     &\textbf{84.79$\pm$0.22}         &83.07$\pm$1.07    \\
		PTC\_MM         &50.82$\pm$6.20     &\textbf{66.68$\pm$5.71}      &62.18$\pm$2.22   &\textbf{66.30$\pm$4.87}     &67.18$\pm$1.62         &\textbf{69.59$\pm$7.39}       \\
		PTC\_MR         &49.68$\pm$2.03      &\textbf{63.38$\pm$6.04}    &59.88$\pm$2.02      &\textbf{67.73$\pm$6.61}    &61.32$\pm$0.89         &\textbf{63.59$\pm$5.31}   \\
		PTC\_FM         &51.94$\pm$4.05        &\textbf{62.83$\pm$6.23}   &61.38$\pm$1.66    &\textbf{64.45$\pm$5.04}   &64.44$\pm$2.09         &\textbf{65.16$\pm$5.62}    \\
		PTC\_FR          &49.54$\pm$6.00      &\textbf{65.82$\pm$1.07}    &66.91$\pm$1.46    &\textbf{68.39$\pm$3.57}   &66.17$\pm$1.02         &\textbf{67.82$\pm$5.03}     \\
		ENZYMES       &23.88$\pm$1.78     &\textbf{30.50$\pm$3.88}    &41.07$\pm$0.77    &\textbf{50.33$\pm$4.70}    &51.98$\pm$1.24        &\textbf{54.33$\pm$6.11}     \\ 
		PROTEINS      &71.44$\pm$0.25        &\textbf{73.77$\pm$2.33}    &75.77$\pm$0.58    &\textbf{76.19$\pm$2.91}     &75.45$\pm$0.20         &\textbf{75.47$\pm$3.26}    \\
		IMDB-BINARY &67.03$\pm$0.79       &\textbf{69.60$\pm$4.80}   &72.20$\pm$0.78    &\textbf{74.60$\pm$4.74}    &72.26$\pm$0.78         &\textbf{78.10$\pm$5.26}    \\
		IMDB-MULTI  &40.83$\pm$0.57        &\textbf{42.80$\pm$2.84}    &\textbf{50.89$\pm$0.90}    &48.33$\pm$2.70     &50.39$\pm$0.49         &\textbf{53.33$\pm$3.89}    \\
		COLLAB          &72.84$\pm$0.28        &\textbf{73.92$\pm$2.03}    &N / A    &N / A     &\textbf{78.90$\pm$1.90}         &75.54$\pm$2.78    \\
		\bottomrule
	\end{tabular}
\end{table*}

 \begin{table*}[!htb]
	\centering
	\caption{Comparison of classification accuracy ($\pm$ standard deviation) of \textsc{DeepMap} to other competitors on the benchmark datasets.}
	\label{tab:classification2}
	\begin{tabular}{l|l|l|l|l|l|l|l|l}
		\toprule
		Dataset             &\textsc{DeepMap}                    &\textsc{DGCNN}          &\textsc{GIN}          &\textsc{DCNN}       &\textsc{PatchySan}            &\textsc{DGK}           &\textsc{RetGK}             &\textsc{GNTK}\\ \hline
		SYNTHIE             &\textbf{54.53$\pm$6.16}     &47.50$\pm$7.99    &53.48$\pm$3.64       &54.18$\pm$4.49     &44.25$\pm$14.36        &52.43$\pm$1.02   &49.95$\pm$1.96      &53.98$\pm$0.87\\
		KKI                   &\textbf{62.92$\pm$7.94}     &56.25$\pm$18.8         &60.34$\pm$12.5      &48.93$\pm$7.50    &43.75$\pm$13.98        &51.25$\pm$4.17    &48.50$\pm$2.99    &46.75$\pm$5.75\\
		BZR\_MD           &\textbf{73.55$\pm$5.76}     &64.67$\pm$9.32    &70.53$\pm$8.00        &59.61$\pm$11.2     &67.00$\pm$9.48        &58.50$\pm$1.52   &62.77$\pm$1.69      &66.47$\pm$1.20\\ 
		COX2\_MD        &\textbf{72.28$\pm$9.37}     &64.00$\pm$8.86    &65.97$\pm$5.70     &51.29$\pm$5.31     &65.33$\pm$7.78         &51.57$\pm$1.71    &59.47$\pm$1.66    &64.27$\pm$1.55\\ 
		DHFR                &\textbf{85.17$\pm$2.19}     &70.67$\pm$4.95     &82.15$\pm$4.02      &59.80$\pm$2.45     &77.00$\pm$3.59         &64.13$\pm$0.89    &82.33$\pm$0.66    &73.48$\pm$0.65\\ 
		NCI1                 &83.07$\pm$1.07                    &71.73$\pm$2.14    &82.70$\pm$1.70    &57.10$\pm$0.69     &78.60$\pm$1.90         &80.31$\pm$0.46   &\textbf{84.50$\pm$0.20} &84.20$\pm$1.50  \\
		PTC\_MM         &\textbf{69.59$\pm$7.39}     &62.12$\pm$14.1      &67.19$\pm$7.41       &63.04$\pm$2.71     &56.58$\pm$9.01         &67.09$\pm$0.49      &67.90$\pm$1.40   &65.94$\pm$1.21  \\
		PTC\_MR         &\textbf{67.73$\pm$6.61}      &55.29$\pm$9.38      &62.57$\pm$5.18       &55.65$\pm$4.92    &55.25$\pm$7.98         &62.03$\pm$1.68   &62.50$\pm$1.60    &58.32$\pm$1.00\\
		PTC\_FM         &\textbf{65.16$\pm$5.62}        &60.29$\pm$6.69      &64.22$\pm$2.36      &63.50$\pm$3.78   &58.38$\pm$9.27         &64.47$\pm$0.76    &63.90$\pm$1.30    &63.85$\pm$1.20\\
		PTC\_FR          &\textbf{68.39$\pm$3.57}      &65.43$\pm$11.3       &66.97$\pm$6.17        &66.24$\pm$3.83   &61.00$\pm$5.61         &67.66$\pm$0.32    &67.80$\pm$1.10    &66.97$\pm$0.56\\
		ENZYMES       &54.33$\pm$6.11                      &43.83$\pm$6.85       &50.50$\pm$6.01       &17.50$\pm$2.67    &22.50$\pm$7.08        &53.43$\pm$0.91 &\textbf{60.40$\pm$0.80}    &32.35$\pm$1.17\\ 
		PROTEINS        &\textbf{76.19$\pm$2.91}        &73.06$\pm$4.81    &\textbf{76.20$\pm$2.80}    &66.47$\pm$1.10     &75.90$\pm$2.80         &75.68$\pm$0.54  &75.80$\pm$0.60 &75.60$\pm$4.20  \\
		IMDB-BINARY &\textbf{78.10$\pm$5.26}       &70.03$\pm$0.86        &75.10$\pm$5.10       &71.38$\pm$2.08    &71.00$\pm$2.29         &66.96$\pm$0.56    &72.30$\pm$0.60     &76.90$\pm$3.60\\
		IMDB-MULTI  &\textbf{53.33$\pm$3.89}       &47.83$\pm$0.85          &52.30$\pm$2.80       &45.02$\pm$1.73     &45.23$\pm$2.84         &44.55$\pm$0.52     &48.70$\pm$0.60    &52.80$\pm$4.60\\
		COLLAB            &75.54$\pm$2.78        &73.76$\pm$2.52    &80.20$\pm$1.90    &76.24$\pm$0.60     &72.60$\pm$2.20         &73.09$\pm$0.25   &81.00$\pm$0.30 &\textbf{83.60$\pm$1.00}   \\
		\bottomrule
	\end{tabular}
\end{table*}

 \begin{table*}[!htb]
	\centering
	\caption{Comparison of classification accuracy ($\pm$ standard deviation) of \textsc{DeepMap} to other GNNs with the same input of vertex feature maps.}
	\label{tab:classification3}
	\begin{tabular}{l|l|l|l|l|l}
		\toprule
		Dataset             &\textsc{DeepMap}                    &\textsc{DGCNN}                &\textsc{GIN}                           &\textsc{DCNN}                     &\textsc{PatchySan}            \\ \hline
		SYNTHIE            &\textbf{54.53$\pm$6.16}      &47.25$\pm$7.86               &53.68$\pm$8.25                  &50.67$\pm$4.41                 &42.00$\pm$10.36        \\
		KKI                      &62.92$\pm$7.94                    &56.25$\pm$18.87             &\textbf{64.93$\pm$17.15}   &53.93$\pm$7.22                 &48.75$\pm$15.26        \\
		BZR\_MD           &\textbf{73.55$\pm$5.76}      &64.33$\pm$8.90                &73.00$\pm$10.70                 &68.73$\pm$3.46                &67.33$\pm$8.41        \\ 
		COX2\_MD        &\textbf{72.28$\pm$9.37}      &59.00$\pm$9.30                &65.76$\pm$7.65                  &61.98$\pm$4.99                 &62.00$\pm$10.13         \\ 
		DHFR                &\textbf{85.17$\pm$2.19}        &79.33$\pm$5.56                &80.16$\pm$5.27                   &76.51$\pm$6.47                  &71.00$\pm$16.76         \\ 
		NCI1                 &\textbf{83.07$\pm$1.07}        &71.05$\pm$2.03                  &75.38$\pm$2.03                   &77.34$\pm$0.98                 &80.14$\pm$1.58          \\
		PTC\_MM         &\textbf{69.59$\pm$7.39}     &61.21$\pm$12.27                 &68.40$\pm$7.78                   &64.64$\pm$2.74                 &62.00$\pm$7.69         \\
		PTC\_MR         &\textbf{67.73$\pm$6.61}      &54.12$\pm$7.74                   &64.87$\pm$8.41                   &57.57$\pm$4.26                  &58.88$\pm$8.19         \\
		PTC\_FM         &\textbf{65.16$\pm$5.62}        &58.53$\pm$6.86                 &61.89$\pm$8.54                  &57.78$\pm$4.07                  &58.38$\pm$5.09         \\
		PTC\_FR          &\textbf{68.39$\pm$3.57}      &65.43$\pm$11.38                 &66.08$\pm$5.99                  &62.99$\pm$4.17                  &58.25$\pm$8.81         \\
		ENZYMES       &\textbf{54.33$\pm$6.11}       &35.33$\pm$5.02                    &37.50$\pm$3.59                    &42.75$\pm$1.81                   &25.17$\pm$5.19        \\ 
		PROTEINS        &76.19$\pm$2.91                   &\textbf{76.58$\pm$4.37}    &75.10$\pm$5.04                     &65.55$\pm$3.36                  &65.50$\pm$6.80           \\
		IMDB-BINARY &\textbf{78.10$\pm$5.26}    &69.20$\pm$5.73                     &74.10$\pm$3.18                    &74.55$\pm$2.50                   &68.70$\pm$5.27         \\
		IMDB-MULTI  &\textbf{53.33$\pm$3.89}     &47.67$\pm$4.41                    &49.87$\pm$3.14                    &48.32$\pm$3.40                   &43.33$\pm$7.25        \\
		COLLAB            &75.54$\pm$2.78                 &73.5$\pm$2.1                         &71.68$\pm$2.10                      &\textbf{76.50$\pm$1.26}     &72.38$\pm$2.18            \\
		\bottomrule
	\end{tabular}
\end{table*}

\subsubsection{Classification Accuracy}
Table \ref{tab:classification1} shows the classification accuracies of the deep map models and their corresponding graph kernels on the benchmark datasets. We can see from the table that the deep map models outperform their corresponding graph kernels in most cases. On the dataset IMDB-MULTI, \textsc{SP} is better than \textsc{DeepMap-SP}, with a gain of 5.3\%. On the dataset NCI1 and COLLAB, \textsc{WL} outperforms \textsc{DeepMap-WL}. However, on the other datasets, e.g., BZR\_MD, \textsc{DeepMap-GK} has a gain of 28.1\% over \textsc{GK}, \textsc{DeepMap-SP} has a gain of 7.2\% over \textsc{SP}, and \textsc{DeepMap-WL} has a gain of 19.9\% over \textsc{WL}, respectively.

Table \ref{tab:classification2} shows the classification accuracies of \textsc{DeepMap} and other graph kernels and GNNs on the benchmark datasets. \textsc{DeepMap} outperforms all the GNNs on most of the datasets. \textsc{DGK} is also based on the graph feature maps. We can see that our model \textsc{DeepMap} is superior to \textsc{DGK} with a large margin on all the datasets. On the dataset ENZYMES and NCI1, \textsc{RetGK} outperforms \textsc{DeepMap}. On the dataset COLLAB, \textsc{GNTK} is better than \textsc{DeepMap}. However, \textsc{DeepMap} dramatically outperforms the worst method \textsc{DCNN} with a gain of 210.5\%. On the dataset COX2\_MD, \textsc{DeepMap} has a gain of 9.6\% over the second-best method \textsc{GIN} and has a gain of 40.9\% over the worst method \textsc{DCNN}. 

In the next experiment, we input the vertex feature maps to other GNNs. Table \ref{tab:classification3} shows the classification accuracies. We want to investigate if \textsc{DeepMap} has a better architecture for vertex feature maps. Even with the same inputs as \textsc{DeepMap}, all the other GNNs cannot defeat \textsc{DeepMap} in most cases. On the dataset KKI, \textsc{GIN} achieves the best classification result, with a gain of 3.2\% over \textsc{DeepMap}. On the dataset PROTEINS, \textsc{DGCNN} is slightly better than \textsc{DeepMap}. On the dataset COLLAB, \textsc{DCNN} achieves the best result.

\subsection{Runtime}
Table~\ref{tab:runtime} shows the average runtime of each epoch of \textsc{DeepMap} and other GNNs on the real-world datasets. \textsc{DeepMap} is competitive to other GNNs. PTC\_MM, PTC\_MR, PTC\_FM and PTC\_FR are four similar datasets. The runtime of \textsc{DeepMap} on these four datasets are differing because \textsc{DeepMap} uses different kinds of vertex feature maps and their dimensions are different. For datasets NCI1, ENZYMES, IMDB-BINARY and IMDB-MULTI, \textsc{DeepMap} performs the worst because the vertex feature maps built on the shortest-path or subtree patterns are of high dimension. Each epoch of \textsc{GIN} costs over 1s because \textsc{GIN} uses five layers of MLPs (multilayer perceptrons) that are hard to train.

 \begin{table*}[!htb]
	\centering
	\caption{Runtime of each epoch of \textsc{DeepMap} and other GNNs.}
	\label{tab:runtime}
	\begin{tabular}{l|l|l|l|l|l}
		\toprule
		Dataset             &\textsc{DeepMap}     &\textsc{DGCNN}    &\textsc{GIN}    &\textsc{DCNN}    &\textsc{PatchySan}            \\ \hline
		SYNTHIE            &166.7ms                    &313.5ms                  &1.4s                  &338.5ms               &566.0ms        \\
		KKI                      &428.8ms                    &61.5ms                   &1.1s                    &63.1ms                 &343.9ms        \\
		BZR\_MD           &99.2ms                       &224.0ms                &1.1s                    &93.3ms                &366.0ms        \\ 
		COX2\_MD        &106.9ms                      &200.5ms                &1.2s                   &95.0ms                &367.8ms         \\ 
		DHFR                  &564.2ms                     &442.5ms                &1.2s                   &375.8ms              &654.1ms         \\ 
		NCI1                    &7.3s                              &3.0s                        &1.6s                   &3.4s                     &2.5s          \\
		PTC\_MM           &104.3ms                      &212.5ms                 &1.1s                    &138.3ms              &381.2ms         \\
		PTC\_MR            &213.0ms                      &212.5ms                 &1.1s                    &148.1ms              &390.5ms         \\
		PTC\_FM            &430.3ms                      &217.5ms                 &1.1s                   &147.2ms               &382.9ms         \\
		PTC\_FR             &121.1ms                        &219.5ms                &1.1s                   &143.8ms               &385.0ms         \\
		ENZYMES          &9.9s                               &359.5ms                &1.2s                 &279.1ms                &530.6ms        \\ 
		PROTEINS         &334.1ms                        &727.5ms                &1.2s                  &1.2s                       &887.2ms           \\
		IMDB-BINARY    &2.9s                              &638.0ms                &1.2s                  &514.0ms               &932.8ms         \\
		IMDB-MULTI      &2.6s                              &882.0ms                &1.3s                  &665.7ms              &1.1s        \\
		COLLAB              &8.4s                              &6.3s                        &3.8s                  &10.4s                    &4.1s            \\
		\bottomrule
	\end{tabular}
\end{table*}


\section{Discussion}
Similar to \textsc{PatchySan}, our method \textsc{DeepMap} also imposes an order for graph vertices to make alignments across graphs. However, \textsc{DeepMap} is different from \textsc{PatchySan} in three aspects:  (1) \textsc{DeepMap} adopts eigenvector centrality to impose an order for graph vertices, which is more efficient than \textsc{Nauty} used in \textsc{PatchySan}. (2) \textsc{PatchySan} samples a number (equals to the average degree) of vertices from graphs to construct a vertex sequence. \textsc{DeepMap} uses all the vertices in a graph to generate a vertex sequence. Compared with \textsc{PatchySan}, \textsc{DeepMap} makes full use of all the vertex information in a graph. (3) The input to \textsc{PatchySan} is the one-hot encoding of each vertex label, while the input to \textsc{DeepMap} is the vertex feature map built on the graphlet, shortest-path, or subtree patterns. Compared with the one-hot encoding of each vertex label, vertex feature maps include richer information. The disadvantage of using the vertex feature map is that the dimension may be very high and it leads to low efficiency for CNNs. 

As discussed in Section~\ref{relatedWork}, DeepTrend 2.0 maps a sensor network to an image. Neighboring sensors may not be mapped to neighboring pixels. Differing from DeepTrend 2.0, \textsc{DeepMap} maps a graph into a vertex sequence. Then, for each vertex in the vertex sequence, \textsc{DeepMap} decides its receptive field using BFS on the original graph. All the vertices in a receptive field are neighboring vertices in the original graph. One problem with this formalization is that the size of the input vertex sequence into CNNs is $r$ times that of the original graph. This may also cause the low-efficiency problem. \textsc{DeepMap} uses a summation layer as a readout function for the whole graph. The sum function loses the local distribution of each deep vertex feature map. A possible alternative is to use a concatenation layer that concatenates all the deep vertex feature maps into a vector. \textsc{DeepMap} is built on the hand-crafted vertex feature maps used in graph kernels. It is not an end-to-end framework. Recently, researchers have been focusing on deriving neural architectures from graph kernels~\cite{lei2017deriving}. It is very interesting to research on this direction, designing an end-to-end neural learning architecture that is inspired from the mechanisms of graph kernels. Another interesting direction is to design a new graph neural network that can realize different levels of embeddings, including node level, edge level, group level and graph level.

As discussed before, R-convolution graph kernels just decompose graphs into substructures and compare these substructures. Thus, they cannot capture the high-order complex interactions between vertices. For example, the random walk graph kernel~\cite{gartner2003graph,vishwanathan2010graph,kashima2003marginalized} conduct random walk on each vertex in two graphs. Each random walk is denoted as a string of node labels and edge labels. Then, the random walk graph kernels just count the number of common random walks (the same strings of node labels and edge labels) in these two graphs. Because random walk is conducted on the first-order transition matrix of the graph structure, the random walk graph kernel cannot capture the high-order complex interactions between vertices. To this end, one possible extension is to conduct random walk on the high-order transition matrix of the graph structure. We leave this for a future work.

\section{Conclusion}\label{conclusion}
In this paper, we have proposed the deep map models to learn deep representations for graphs. \textsc{DeepMap} extends CNNs from images to graphs of arbitrary shape and size, by solving the problems of vertex alignment across graphs and vertex receptive field generation. \textsc{DeepMap} can be built on the vertex feature maps of any substructures. By resolving the two main problems that derived from R-convolutional graph kernels, \textsc{DeepMap} dramatically improves the performances of  R-convolutional graph kernels and also outperforms several state-of-the-art graph neural networks. The learned deep feature map of each vertex can also be considered as vertex embedding and used for vertex classification. In the future, we would like to develop new architectures that integrate the mechanisms of graph kernels for graph neural networks.

\section*{Acknowledgment}
The authors would like to thank anonymous reviewers for their constructive and helpful comments. This work was supported partially by the National Science Foundation (grant \# IIS-1817046) and by the U.S. Army Research Laboratory and the U.S. Army Research Office (grant \# W911NF-15-1-0577).

\bibliographystyle{IEEEtran}
\bibliography{reference}

\end{document}